%% file: JSSP.tex
\documentclass[final,3p]{elsarticle}

\usepackage{multirow}
\usepackage{multicol}
\usepackage{array}
\usepackage{bm}
\usepackage{amsmath}
\usepackage{amsfonts}

\usepackage{booktabs}

\usepackage{amssymb}
\usepackage{lineno}
\usepackage{subfig}
\usepackage{algorithm}
\usepackage[noend]{algpseudocode}
\usepackage{multirow}
\usepackage{booktabs}
\usepackage{multicol}
\usepackage{array}
\usepackage{bm}
\usepackage{amsmath}

\usepackage{hyperref}

\usepackage{graphicx}

\usepackage{tikz}
\usepackage{pgfplots}
\usetikzlibrary{patterns}
\usepackage{csvsimple}
\usepackage{pgfplotstable}

\usepackage[final,markup=none,commentmarkup=footnote]{changes}
\usepackage{color}
\definechangesauthor[color=magenta]{AE}
\definechangesauthor[color=blue]{YS}
\newcommand{\note}[2][]{\added[id=#1,comment={\color{authorcolor}#2}]{}}

\graphicspath{{figs/}}

\definecolor{olivegreen}{RGB}{85,107,47}
\definecolor{orange}{RGB}{229,148,0}
\definecolor{marine}{RGB}{0,32,96}
\definecolor{maroon}{RGB}{178, 50, 50}

\usepackage{xcolor,colortbl}
\definecolor{Gray}{gray}{0.90}
\definecolor{LightCyan}{rgb}{0.88,1,1}
\definecolor{wheat}{rgb}{0.96, 0.87, 0.7}
\definecolor{teagreen}{rgb}{0.82, 0.94, 0.75}

\newcolumntype{g}{>{\columncolor{Gray}}r}
\newcolumntype{o}{>{\columncolor{orange}}r}
\newcolumntype{s}{>{\columncolor{LightCyan}}r}
\newcolumntype{w}{>{\columncolor{wheat}}r}
\newcolumntype{t}{>{\columncolor{teagreen}}r}

\pgfplotsset{every tick label/.append style={font=\tiny}}
\pgfplotsset{
	box plot width/.initial=4em,
	box plot/.style={
		/pgfplots/.cd,
		black,
		only marks,
		mark=-,
		mark size=\pgfkeysvalueof{/pgfplots/box plot width},
		/pgfplots/error bars/.cd,
		y dir=plus,
		y explicit,
	},
	box plot box/.style={
		/pgfplots/error bars/draw error bar/.code 2 args={%
			\draw [line width=0.20mm]  ##1 -- ++(\pgfkeysvalueof{/pgfplots/box plot width},0pt) |- ##2 -- ++(-\pgfkeysvalueof{/pgfplots/box plot width},0pt) |- ##1 -- cycle;
		},
		/pgfplots/table/.cd,
		y index=2,
		y error expr={\thisrowno{3}-\thisrowno{2}},
		/pgfplots/box plot
	},
	box plot top whisker/.style={
		/pgfplots/error bars/draw error bar/.code 2 args={%
			\pgfkeysgetvalue{/pgfplots/error bars/error mark}%
			{\pgfplotserrorbarsmark}%
			\pgfkeysgetvalue{/pgfplots/error bars/error mark options}%
			{\pgfplotserrorbarsmarkopts}%
			\path ##1 -- ##2;
		},
		/pgfplots/table/.cd,
		y index=4,
		y error expr={\thisrowno{2}-\thisrowno{4}},
		/pgfplots/box plot
	},
	box plot bottom whisker/.style={
		/pgfplots/error bars/draw error bar/.code 2 args={%
			\pgfkeysgetvalue{/pgfplots/error bars/error mark}%
			{\pgfplotserrorbarsmark}%
			\pgfkeysgetvalue{/pgfplots/error bars/error mark options}%
			{\pgfplotserrorbarsmarkopts}%
			\path ##1 -- ##2;
		},
		/pgfplots/table/.cd,
		y index=5,
		y error expr={\thisrowno{3}-\thisrowno{5}},
		/pgfplots/box plot
	},
	box plot median/.style={
		/pgfplots/box plot
	}
}

\begin{document}

\begin{frontmatter}
	
	\title{Enhancing Constraint Programming via Supervised Learning \\ for Job Shop Scheduling}
	
	\author[label1]{Yuan~Sun\corref{cor1}}
	\ead{yuan.sun@latrobe.edu.au}
	\cortext[cor1]{Corresponding author}
	\author[label1]{Su~Nguyen}
	\ead{p.nguyen4@latrobe.edu.au}
	\author[label2]{Dhananjay~Thiruvady}
	\ead{dhananjay.thiruvady@deakin.edu.au}
	\author[label3]{Xiaodong Li}
	\ead{xiaodong.li@rmit.edu.au}
	\author[label4]{Andreas~T.~Ernst}
	\ead{andreas.ernst@monash.edu}
	\author[label5]{Uwe~Aickelin}
	\ead{uwe.aickelin@unimelb.edu.au}

	\address[label1]{La Trobe Business School, La Trobe University, Bundoora, VIC, 3086, AU.}	
	\address[label2]{School of Information Technology, Deakin University, Geelong, AU.}
	\address[label3]{School of Computing Technologies, RMIT University, Melbourne, VIC, 3000,  AU}	
	\address[label4]{School of Mathematics, Monash University, Clayton, VIC, 3800,  AU}
	\address[label5]{School of Computing and Information Systems, The University of Melbourne, Parkville, VIC, 3010, AU}	

\begin{abstract}

Constraint programming (CP) is a powerful technique for solving constraint satisfaction and optimization problems. In CP solvers, the variable ordering strategy used to select which variable to explore first in the solving process has a significant impact on solver effectiveness. To address this issue, we propose a novel variable ordering strategy based on supervised learning, which we evaluate in the context of job shop scheduling problems. Our learning-based methods predict the optimal solution of a problem instance and use the predicted solution to order variables for CP solvers. \added[]{Unlike traditional variable ordering methods, our methods can learn from the characteristics of each problem instance and customize the variable ordering strategy accordingly, leading to improved solver performance.} Our experiments demonstrate that training machine learning models is highly efficient and can achieve high accuracy. Furthermore, our learned variable ordering methods perform competitively when compared to four existing methods. Finally, we demonstrate that hybridising the machine learning-based variable ordering methods with traditional domain-based methods is beneficial.

\end{abstract}

\begin{keyword}
	Constraint programming, supervised learning, variable ordering, job shop scheduling problem. 
\end{keyword}

\end{frontmatter}

\input{01-introduction}
\input{02-background}
\input{03-method}
\input{04-experiments}
% \input{05-related_work}
\input{06-conclusion}
\bibliographystyle{IEEEtranNEmph}
\bibliography{JSSP,gpsurvey}

\vfill

\end{document}

%% file: 01-introduction.tex
\section{Introduction}\label{sec:introduction}

Constraint programming (CP) is a powerful technique for solving both constraint satisfaction and optimization problems \cite{apt2003principles}. The efficiency of CP solvers is substantially impacted by variable ordering, which determines the order in which variables are branched during the search process \cite{gent1996empirical}. An effective variable ordering strategy should result in a smaller search tree, thus enabling more efficient problem solving. However, there is currently a lack of theoretical understanding regarding what constitutes a good variable ordering, and empirically discovering an effective strategy can be challenging \cite{liberatore2000complexity}.

Most existing variable ordering strategies are based on the intuition of human experts \cite{haralick1980increasing,bessiere1996mac,refalo2004impact,michel2012activity,smith1997trying,wattez2019refining}. However, recent advances in deep reinforcement learning have enabled the automatic learning of variable ordering strategies for solving constraint satisfaction problems, using a graph neural network \cite{song2019learning}. Additionally, a multi-armed bandits approach was proposed in \cite{wattez2020learning} to select a suitable variable ordering strategy from a set of existing ones for a given instance. While these methods were primarily designed for constraint \emph{satisfaction} problems, which focus on finding a feasible solution or proving that none exists, a more recent study \cite{nguyen2021genetic} has introduced a genetic programming approach to evolve variable ordering strategies for solving constraint \emph{optimization} problems. This evolutionary learning method aims to learn a variable ordering that minimizes the number of branches required by a CP solver to optimally solve a set of training instances.

\added[]{Inspired by these recent developments, we propose a \emph{supervised} learning approach to automatically learn a variable ordering strategy for a CP solver. Our approach is motivated by the idea that ordering variables based on the optimal solution of a problem instance can lead to high-quality solutions early in the search process and enable aggressive search space pruning. To achieve this, we develop a supervised learning model that predicts the optimal solution for a given problem instance, which we then use to order variables for more efficient problem-solving. Unlike traditional variable ordering methods, our approach can learn from the characteristics of each problem instance and customize the variable ordering strategy accordingly, resulting in improved solver performance. Furthermore, compared to existing learning-based methods, our approach based on optimal solution prediction is more interpretable and easier to train.}

\added[]{In this study, we focus on evaluating the efficacy of our proposed method on a class of NP-hard problems called job shop scheduling (JSS) problems \cite{pinedo_scheduling:_2008}. JSS problems are prevalent in manufacturing, where the goal is to schedule the operations of multiple jobs on several machines subject to constraints while optimizing an objective function such as makespan or total weighted tardiness.  In this context, we demonstrate that optimal solution prediction for JSS can be modeled as either a regression or classification task. The regression task aims to predict the optimal start times of operations based on the extracted features, while the classification task predicts, for each pair of operations in a JSS instance, which operation should be scheduled earlier in the optimal solution.}

We utilize solved JSS instances with known optimal solutions to train our models using several machine learning (ML) algorithms, including support vector machine, multi-layer perceptron, and genetic programming. Our experimental results demonstrate that our supervised learning models achieve high accuracy in predicting optimal solutions for JSS, and they are highly efficient to train. When integrated into a CP solver for JSS, our learned variable ordering methods outperform three traditional methods and one state-of-the-art learning-based approach. Additionally, we show that combining our ML-based variable ordering methods with a traditional domain-based method leads to even better performance. 

\noindent The contributions of this paper can be summarised as follows:
\begin{enumerate}
    \item We propose two novel supervised learning models (regression and classification) that predict optimal solutions for JSS with high accuracy.
    \item We introduce a new automatic variable ordering method based on the predicted optimal solutions to enhance CP solvers.
    \item We conduct an extensive experimental study to evaluate the efficacy of the proposed variable ordering methods and demonstrate their competitive performance compared to state-of-the-art methods.
    \item We show the benefits of combining ML-based and domain-based variable ordering methods through hybridisation.
\end{enumerate}

\added[]{The structure of this paper is organised as follows. In Section~\ref{sec:problem}, we provide the background and related work. Section~\ref{sec::methods} describes our proposed variable ordering methods based on supervised learning models. We present and analyse our experimental results in Section~\ref{sec::experiments}. Finally, in Section~\ref{sec:conclusion}, we conclude the paper by summarizing our contributions and discussing potential future research directions.}

%% file: 02-background.tex
\section{Background and Related Work} \label{sec:problem}

\subsection{Job shop scheduling}

The JSS problem requires a number of jobs to be processed on several machines such that an objective function is optimised. In this context, a \emph{job} consists of a number of operations, and each operation must be executed on a specific machine. In JSS, the execution of operations on machines (routes) are fixed and can be different for different jobs  \cite{pinedo_scheduling:_2008}. In this study, the focus is on the static JSS problem, where a schedule for the \emph{shop} (the working/manufacturing environment) must be determined, with the shop consisting of a set of $M$ machines and $N$ jobs. A job $j$ has a fixed route through the machines, and the processing times are different at each machine it visits. In the notation of \cite{brucker2007classification} this scheduling problem is $J | r_j | Obj$ where $Obj$ is one of $C_{\max}$, $T_{\max}$ or $\sum w_jT_j$.

The following notation is used to define the JSS. A number of jobs $\mathcal{J} = \{1, \dots, j, \dots, N\}$ are given, and job $j$ has associated with it a release time $r_j$, a due date $d_j$, a weight $w_j$ and operations $\{o_{j1}, \dots, o_{jn_j}\}$, where $n_j$ is the number of operations of job $j$. The processing times of all operations of job $j$ are $time_j=(p_{j1}, \dots, p_{jn_j})$, where $p_{ji}$ is the processing time of job $j$'s $i^{\mathrm{th}}$ operation. Moreover, the route for job $j$ is $route_j=(m_{j1}, \dots, m_{jn_j})$, which specifies the sequence of machines that job $j$ will visit, where $m_{ji}$ is a machine that will process job $j$'s $i^{\mathrm{th}}$ operation.

The following four sets of variables are also defined: 
\begin{itemize}
    \item $s_{ji}$ is the starting time of job $j$'s $i^{\mathrm{th}}$ operation,
    \item $e_{ji}$ is the end time of job $j$'s $i^{\mathrm{th}}$ operation,
    \item $C_j$ is the completion time of job $j$, and 
    \item $T_j$ is the tardiness of job $j$: $T_j=\max(C_j-d_j,0)$.
\end{itemize}

There are three different objectives considered for the JSS problem in this study. These objectives are typically seen in industrial problems and have been used as the most common performance measures in previous studies \cite{nguyen_computational_2013}.
\begin{enumerate}
    \item Makespan: $C_{max} = \max_{j \in \mathcal{J}}\{ C_j \}$,
    \item Maximum tardiness: $T_{max}= \max_{j \in \mathcal{J}}\{T_j\}$,
    \item 	Total weighted tardiness: $TWT = \sum_{j \in \mathcal{J}}(w_j \times T_j)$.
\end{enumerate}

\subsection{Constraint programming for JSS}
In a CP solver, a problem instance is typically represented by a constraint network $NW(X,D,C)$, where $X$ is the set of decision variables, $D$ is the set of domains (i.e., possible values that a variable can take) of the variables, and $C$ is the set of constraints. In the context of JSS, the variables are start and end times of operations, completion time and tardiness of jobs defined before. The set of constraints are  
% The following constraint programming model for the JSS forms the basis for the methods that are tested in this study. 
% We now provide a constraint programming model for the JSS, which forms the basis for the algorithms that are tested in this study.  
% \DT{I believe this model can stay as is. I don't really think this part can be called self plagiarism. Let me know if you think otherwise.}
\begin{align}
\small
    &\forall j \in {\cal J}: & & \quad s_{j1} \geq r_j \label{c1}\\
    &\forall j \in {\cal J},\  i \in \{1,\ldots,n_j\}: &\quad & \quad e_{ji} = s_{ji} + p_{ji} \label{c2} \\
    &\forall j \in {\cal J}:   & & \quad C_{j} = e_{jn_j} \label{c3} \\
    &\forall j \in {\cal J}: &   & \quad T_{j} = \max(C_j - d_j, 0) \label{c4}
\end{align}
Constraint~\eqref{c1} ensures the release times of each job are satisfied. Constraint~\eqref{c2} links the start and end time variables. Constraint~\eqref{c3} sets $C_j$ to be the completion time of the last operation of job $j$, and Constraint~\eqref{c4} computes the tardiness $T_j$ of job $j$. Moreover, there is no overlap between the operations on the same machine, which is implemented using the following disjunctive constraints:
\begin{equation}\small
\begin{split}
    \forall j,k \in {\cal J}, u \in \{1,\ldots,n_j\}, v & \in \{1,\ldots,n_k\}:  \\
     m_{ju} = m_{kv}   \Rightarrow & s_{ju} \geq  e_{kv} \vee s_{kv}  \geq e_{ju}.
\end{split}
\end{equation}

In other words, if two operations $u$ and $v$ of two different jobs need to execute on the same machine (i.e. $m_{ju} = m_{kv}$), the start time of the first job must be later than the end time of the second job, or vice versa. There may also exist precedences between operations of a job:
\begin{equation}\small
    \forall j \in {\cal J}, i \in \{1,\ldots,n_j-1\}: \quad s_{j,i+1} \geq e_{ji}. 
\end{equation}

Finally, each of the three objective functions are implemented in the model as follows:
\begin{enumerate}
    \item Makespan: let $C_{max}$  be the largest completion time across all jobs. In addition to the above model, the following objective and constraint are included:
    \begin{align}\small
        \min. & \quad C_{max}, \\ 
        \forall j \in {\cal J}: & \quad C_{max}  \geq C_j. \label{cons:cmax}    %AE: C_j=e_{jn_j} but this is I think more readable
    \end{align}
    Note that due dates are ignored for the Makespan objective. 
    \item Maximum tardiness:  let $T_{max}$ represent the maximum tardiness across all jobs. The following objective and constraint are added to the above model:
    \begin{align}\small
        \min. & \quad T_{max}, \\
        \forall j \in {\cal J}: & \quad T_{max}  \geq T_j. \label{cons:tmax}    
    \end{align} 
    \item Total weighted tardiness (TWT):  In addition to the above model, the following objective (minimise cumulative tardiness' across all jobs) is included:
    \begin{align}\small
        \min. & \quad \sum_{j \in {\cal J}} w_jT_j. \label{cons:twt}    
    \end{align} 
\end{enumerate}

In solving the model, new restrictions are typically identified and incorporated into the system. The solver uses these new restrictions to automatically update the variable domains and potentially identify new restrictions. Specifically, the solver imposes a restriction by checking the domains of all variables for consistency. An outcome of this is that the solver indicates success if the domains are consistent, and if not, the solver is in a state of failure. If imposing a restriction or a set of restrictions is successful, the domains of all variables are updated and any inferred restrictions are imposed. 

In CP terminology, the assignment of a variable to a value is called a labelling step. If such an assignment is made, it is tested for consistency with the variable's current domain. If the assignment is consistent, it is accepted, which leads to propagation of new constraints. Alternatively, the assignment may lead to an inconsistency, in which case a failure is triggered and the labelling step is rejected \cite{apt_2003}. A key aspect of the search component in CP is the variable ordering strategy that guides the selection of variables for labeling. Devising an effective variable ordering strategy can lead to substantial gains in terms of identifying high-quality regions of the search space and better quality solutions more quickly. The focus of our study is to achieve an improved variable ordering strategy in CP via supervised learning, and we demonstrate this can be effectively achieved using JSS as a case study.

\subsection{Methods for tackling job shop scheduling}\label{sec:methodsjss}

Since the JSS problem is NP-hard, finding optimal or high quality solutions by conventional optimization algorithms still remains a challenge \cite{pinedo_scheduling:_2008}. Hence, a number of  optimization approaches have been proposed to tackle JSS. Commercial optimization solvers, such as CPLEX,  have made significant advances in recent years, though solving large problem instances for JSS is not straightforward and scalability still proves to be a challenge. \citet{KU2016165} investigated the performance of four different mixed integer programming models for the JSS, and they found that good results for moderate sized problems can be achieved via a disjunctive model. Additionally, meta-heuristics have been an active area of research in JSS, where these methods are often able to find near-optimal solutions \cite{LI201693,Kreipl00,jensen_2003}. These methods have drawbacks due to a ``combinatorial explosion'', where for large instances, there is an exponential growth in the solution space. This makes it computationally challenging to navigate the search space (locally or globally) efficiently to determine high-quality solutions. Furthermore, due to the inherent nature of meta-heuristics, problem-domain knowledge (e.g. solution neighbourhood structure, schedule construction heuristics, etc.) is necessary to efficiently design an application specific algorithm. This is a difficult process often requiring a significant amount of trial-and-error.

In recent years, CP has been an effective method for solving satisfaction and optimization problems \cite{apt_2003}. There are a number of problem domains where CP is proven to be effective, and scheduling is one of these  \cite{baptiste2012constraint,barreiro2012europa,grimes2010job,Thiruvady2012}. Moreover, CP has also been shown to be effective on JSS. The studies by \citet{beck2011combining} and \citet{watson2008hybrid}, investigated a CP and tabu search hybrid. Their results showed that this hybrid can generate solutions with low variability and are competitive with local search approaches. \citet{grimes2010job} investigated JSS with setup times, and they showed that enhancing the search component of CP can prove to be very effective for this problem. 

\subsection{ML for combinatorial optimization} \label{sec:mlcombinatorialoptimization}

Leveraging ML for combinatorial optimization (MLCO) has attracted significant attention in recent years~\cite{lodi2017learning,bengio2018machine,mazyavkina2021reinforcement,talbi2021machine}. In their survey, \citet{bengio2018machine} have classified MLCO studies into three categories depending on the role of ML: (1) \textit{end-to-end learning} in which ML is used to directly produce solutions for CO problem instances, (2) \textit{learning to configure algorithm} in which ML is used to efficiently parameterise optimization algorithms (in a broad sense), (3) \textit{ML alongside optimization algorithms} in which ML is used to assist low-level decisions within optimization algorithms. In the operations research and computational intelligence community, MLCO has been investigated extensively and referred to as hyper-heuristics \cite{burke_hyper-heuristics:_2013,nguyen2021automated,zhang2021multitask,zhang2021evolving,zhao2022hyperheuristic}. Unlike traditional search algorithms that attempt to find good solutions in the solution search space, hyper-heuristics emphasise on exploring the heuristic (or algorithm) search space to find a good way to solve problem instances. Hyper-heuristics can be used for \textit{heuristic generation} or \textit{heuristic selection}. In heuristic generation, hyper-heuristics use a search algorithm, such as genetic programming (GP), to generate/discover effective heuristics for a problem subset. Generated heuristics can be used to construct solutions or refine/improve a solution (similar to the above mentioned categories 1 and 2 in \citet{bengio2018machine}). When used for heuristic selection, the goal of hyper-heuristics is to adaptively select low-level heuristics in optimization algorithms to solve problem instances (similar to category 3 in \cite{bengio2018machine}). To avoid confusion, we will use \citet{bengio2018machine}'s categorisation to discuss related work.

Many earlier studies in the MLCO literature examine the methods for \textit{end-to-end learning}. For examples, popular ML algorithms such as decision tree \cite{olafsson_learning_2010}, logistic regression \cite{ingimundardottir_supervised_2011}, artificial neural networks \cite{mouelhi-chibani_training_2010}, and deep reinforcement learning~\cite{park2022scalable} can be applied to learn constructive heuristics for production scheduling problems and showed that their learned heuristics outperform existing constructive heuristics in the literature. GP is also a popular learning algorithms in this category and many program representations and advanced search techniques have been proposed to improve the quality of learned heuristics \cite{jakobovic_dynamic_2006,Nguyen2017}. There are several reasons for the popularity of \textit{end-to-end learning}. The main reason is that computational efforts required for training in this case is low and learned model/heuristics can be used to generate good solutions very efficiently. The disadvantage of \textit{end-to-end learning} is that there is no guarantee that solutions produced by these methods are optimal or near-optimal. Empirical results \cite{olafsson_learning_2010,Nguyen2017} showed that there is a gap in solution quality between learned constructive heuristics and optimization algorithms, which is less attractive for problems in which optimal solutions are desirable. Moreover, a meta-algorithm or a heuristic template is usually required to construct solutions, which restrict the applicability of learned heuristics. Apart from production scheduling, \textit{end-to-end learning} is also applied to other CO problems such as NASA's resource-constrained scheduling problem \cite{zhang2000solving}, bin packing \cite{Sim:2015:LLH:2811901.2811903}, graph-based problems \cite{khalil2017learning}, and routing problems \cite{bello17,vanLon2018,yu2019,kool2018attention}. 

In the last decade, there has been a growing interest in \textit{learning to configure algorithm}. For example, \citet{khalil2016learning,balcan2018learning} developed innovative ML techniques to learn variable selection methods for mixed integer linear program (MILP) solvers. \citet{he2014learning,furian2021machine} used ML to learn node selection methods for MILP solvers. \citet{shen2021learning} developed an ML-based primal heuristic for MILP solvers. \citet{nguyen2021genetic} investigated the use of GP to learn variable selectors in CP, and they show that this method is effective for JSS considering three different objectives. Our proposed methods belong to this category because the learned variable selectors will configure the search of CP based on characteristics of problem instances.

% (i.e. \textit{ML alongside optimization algorithms})

With the recent boost in computing power, many innovative ML techniques have been developed to enhance optimization algorithms for solving combinatorial optimization problems. \citet{kruber2017learning} used ML to predict if decomposition is needed to reduce the time to solve MILP instances using features including instance characteristics and decomposition statistics. \citet{Bonami18} used ML to predict if linearisation will be more efficient in solving quadratic programming problem. \citet{basso2020random} used random sampling and ML to learn what are good decompositions \cite{basso2020random}. \citet{Morabit2020mlcs} developed an ML model to select promising columns for column generation. \citet{shen2022enhancing} developed an ML-based pricing heuristic to enhance column generation. \citet{sun2019using,lauri2019fine,sun2020generalization} used ML to prune the search space of combinatorial optimization problems. \citet{sun2022boosting} developed ML models to enhance the sampling of a metaheuristic. 

While the research in this paper fits into this wider effort to use ML to improve combinatorial optimization, it is to the best of our knowledge the first time that it has been used to improve the labelling strategy in CP for solving constraint optimization problems.

%% file: 03-method.tex
\section{Methods}
\label{sec::methods}
We develop two supervised learning models to predict an optimal solution for the JSS problem, which is then used to enhance variable ordering for CP to solve the problem. 
\begin{table*}[!t]
    \centering
    \caption{The features extracted to characterise an operation for JSS. Considering the $i^\mathrm{th}$ operation of job $j$ ($o_{ji}$) to be processed on machine $m_{ji}$, the features for $o_{ji}$ can be computed as below.}
    \label{tab:features}
    \resizebox{\textwidth}{!}{
    \begin{tabular}{lllll}
    \toprule
         & ID & Acronym & Description & Definition  \\\midrule
         & $f_1$ & NPO & the number of precedent operations in job $j$ that need to be processed before $o_{ji}$ & $f_1(o_{ji}) = i - 1 $ \\ 
         & $f_2$ & PT &  the processing time of the operation $o_{ji}$ &  $f_2(o_{ji}) = p_{ji}$ \\
         & $f_3$ & PTB &  the processing time of operations in job $j$ that need to be processed before $o_{ji}$ & $f_3(o_{ji}) = \sum_{k=1}^{i-1}p_{jk}$ \\ 
         & $f_4$ & PTA &  the processing time of operations in job $j$ that need to be processed after $o_{ji}$ & $f_4(o_{ji}) = \sum_{k=i+1}^{n_j}p_{jk}$ \\ 
         & $f_5$ & TPT & the total processing time of the operations in job $j$ & $f_5(o_{ji}) = \sum_{k=1}^{n_j}p_{jk}$ \\ 
         & $f_6$ & DD & the due date of job $j$ that $o_{ji}$ belongs to & $f_6(o_{ji}) = d_j$ \\ 
         & $f_7$ & W & the weight of job $j$ that $o_{ji}$ belongs to &  $f_7(o_{ji}) = w_j$ \\ 
         & $f_8$ & RT & the release time of job $j$ that $o_{ji}$ belongs to & $f_8(o_{ji}) = r_j$ \\ 
         & $f_9$ & EST & the earliest start time of the operation $o_{ji}$ &   $f_9(o_{ji}) = r_j + \sum_{k=1}^{i-1}p_{jk}$ \note[AE]{Are the earliest start time calculated without taking precedences into account? Also is there any reason why the latest start time (that would allow completion without tardiness) isn't considered? No real need to add that at this point though.}\\ 
         & $f_{10}$ & WL & the workload of the machine ($m_{ji}$) that processes the operation $o_{ji}$ & $f_{10}(o_{ji}) = \sum_{u=1}^{N} (p_{uv} \, | \, m_{uv} = m_{ji})$ \\\bottomrule 
    \end{tabular}
    }
\end{table*}

\subsection{Feature extraction} \label{subsec::features}
A key task of building ML models is feature extraction, which often has a significant impact on the accuracy of the models. To predict the optimal value of decision variables (i.e., start times of operations) for JSS, we extract a set of features ($\mathbf{f}$) to characterise an operation in Table~\ref{tab:features}. These features are basic attributes that can be extracted easily from the JSS problem. NPO, PT, PTB, PTA and EST are operation-wise features that characterise an operation, the processing time, downstream and upstream operations etc. In contrast, TPT, DD, W, and RT are job-related features, and WL is a machine-related feature. Although deep learning models such as graph neural networks \cite{scarselli2008graph,kipf2016semi} have the potential to automatically extract features, training deep learning models typically requires a large amount of data and computational resources. Hence, we have only manually designed features to build a simple ML model. However, we will show in experiments that our simple model with little effort of training can already bring significant benefits to CP. 

\subsection{A regression model}
A direct approach to predict optimal solution values of decision variables (i.e., optimal start times of operations) for the JSS problem is to train a regression model based on solved problem instances with known optimal solutions. The regression model takes the features of an operation as inputs, and outputs a real value as the predicted start time for the operation. The training of the regression model is to minimize the error between the predicted and the true optimal start times over a set of training instances. 

To construct a training set, we use the CP-SAT solver in OR-Tools to solve a set of small JSS problem instances to optimality and obtain the optimal solutions. As the problem instances are small, they can be solved very efficiently. An operation in a training problem instance is then represented by a set of features (Table~\ref{tab:features}) and labeled as its start time in the optimal solution. We can then construct a training set with a training example corresponding to an operation in a problem instance. In order to build a more robust regression model, we normalise the training data as follows: The number of precedent operations (NPO) is normalised by the total number of operations in a job; the weight (W) of a job is normalised by the maximum weight of the jobs; the other seven features and the label are normalised by the maximum TPT (total processing time of operations in a job) in a problem instance.

After the training set is obtained, we can train an existing ML algorithm for the regression task. We will evaluate three different ML algorithms, linear support vector machine (SVM), multi-layer perceptron (MLP) and genetic programming (GP). The linear SVM is very fast to train and highly interpretable, as it is a linear model. MLP is a widely-used neural network, that is able to learn complex non-linear relations between input and output data. GP is used as an evolutionary symbolic regressor here. Although it is much slower than other algorithms to train, GP can not only learn complex relations in data but also may have better interpretability. We will use the implementations of SVM and MLP in the Python \emph{scikit-learn} library \cite{scikit-learn} and the GP implementation in the \emph{gplearn} library. \added[]{It is important to note that we have considered multiple ML algorithms to demonstrate the generality of our framework. However, it is beyond the scope of this paper to conduct a comprehensive evaluation of all existing ML algorithms or select the best ML algorithm.}

Given an unseen JSS instance, we can first compute features for each operation, and then use the trained regression model to predict the optimal start time of each operation. 

\subsection{A classification model}
An alternative approach of solution prediction is to train a classification model to predict the optimal sequence of operations for a JSS problem instance. The aim of the classification model is to predict, for each pair of operations in a JSS instance, which operation should be scheduled earlier. It is important to note that this classification model is effectively equivalent to the regression model, in the sense that they would produce the same variable ordering if they are both 100\% accurate in prediction. 

The time complexity of the classification model is quadratic in terms
of the number of operations in a problem instance. To reduce the
computational time, we introduce a variant of the classification model
that only predicts, for each pair of operations \emph{on a machine},
which operation should be scheduled earlier. We observed in our
preliminary experiments that classifying operations within a machine
is easier than classifying operations across different
machines. Note that ordering all of the operations within each machine is sufficient to specify a leaf of the search tree. Hence, we will only describe this classification model variant in detail.  

The training set for the classification model is slightly different from that for the regression model. Here, a training example corresponds to a pair of operations to be processed \emph{on a machine}. Consider a pair of operations, $o_i$ and $o_j$, and their feature vectors, $\mathbf{f}^i$ and $\mathbf{f}^j$, normalised similarly as before. The features of the corresponding training example is the difference between $\mathbf{f}^i$ and $\mathbf{f}^j$: $\mathbf{f}^i - \mathbf{f}^j$. The class label of the training example is $-1$ if the operation $o_i$ is scheduled before $o_j$ in the optimal solution; otherwise $1$. This then becomes a standard binary classification task, and we will test three different ML algorithms for this task: SVM, MLP and GP. \added[]{The reason for selecting these three ML algorithms is the same as before.} 

Given an unseen JSS problem instance, we can apply the trained classification model to predict a sequence of operations. Specifically, we perform pairwise comparison between operations to be processed on a machine, and use the trained classification model to classify which operation is expected to be scheduled earlier on the machine. An operation accumulates a score of 1 if it is predicted to be scheduled later than the other operation under comparison. After the pairwise comparison, each operation has an accumulated score, indicating the predicted order of operations to be processed on a machine. The operations are then sorted in ascending order based on their accumulated scores. 

\subsection{Enhancing variable ordering for CP}

We explore two different ways of enhancing variable ordering for CP via the supervised learning models: (1) a pure ML-based approach, and (2) a hybrid approach. The first approach directly orders variables (operations) based on the predicted solution (i.e., the predicted start times if using the regression model or the predicted operation sequence if using the classification model). On the other hand, the hybrid approach combines the merits of both ML-based with domain-based variable ordering methods. Specifically, we first apply a domain-based variable ordering method, that orders variables based on their minimum domain value. If there are multiple variables with the same minimum domain value, we further use the ML-based variable ordering as a second criterion to break ties. The motivation behind the hybridisation approach is that the domain-based variable ordering method determines predominantly the efficiency of CP for finding feasible solutions \cite{nguyen2021genetic}, whilst our ML-based variable ordering focuses more on finding high-quality solutions.

%% file: 04-experiments.tex
\section{Experiments}\label{sec::experiments}

\subsection{Experimental setup}

\textbf{Problem instances.} We use both randomly generated instances and benchmark instances to evaluate our methods. The random instances are generated using the generator described in \cite{nguyen2021genetic}. For a problem size $N\times M$, we generate 100 problem instances using different random seeds, with the due date allowance factor $h$ set to 1.3. Additionally, we utilize five widely used benchmark datasets from the OR-library \cite{beasley1990or}. Since these datasets lack release times, they will only be applied for the purpose of Cmax.

\textbf{Implementation and platform.} We use the state-of-the-art CP-SAT solver of Google OR-Tools \cite{ortools} as our CP solver. The domain reduction strategy for CP-SAT is set to `selecting-min-value' based on preliminary experiments. The implementations of the ML algorithms are from the \emph{scikit-learn} \cite{scikit-learn} and \emph{gplearn} libraries. The parameters of the ML algorithms are consistent with the default setting except the number of generations for GP is set to 2000. Our source codes are written in Python and will be made publicly available if the paper gets accepted. Our experiments are conducted on a high performance computing server with multiple CPUs @2.70GHz, each with 4GB memory.

\textbf{Baselines.} We compare our ML-based variable ordering methods with four existing methods: (1) \emph{Default}, the default variable ordering method of OR-Tools, that orders variables based on a definition of impact (reduction of search space) \cite{refalo2004impact}, (2) \emph{MinDom}, that orders variables based on the current domain size \cite{haralick1980increasing}, (3) \emph{LowMin}, that orders variables based on the current minimum domain value, and (4) \emph{BranGP}, that uses GP to learn a variable ordering to minimize the number of branches required to solve a set of training instances \cite{nguyen2021genetic}. \added[]{It is worth noting that our comparison has focused solely on various variable ordering methods within CP. We have deliberately avoided comparing our variable ordering method incorporated into CP with other heuristics or meta-heuristics. This is because such a comparison has already been carried out by the \emph{BranGP} method~\cite{nguyen2021genetic}, which we have included in our comparison.}

\begin{table*}[!t]
\centering
\caption{The time (in minutes) taken to collect the training data (including the time used to solve the training problem instances) and to train each variable ordering method.}
\label{tab::training time}
\begin{tabular}{@{\extracolsep{6pt}}lrrrrrrrr@{}}
\toprule
\multirow{2}{*}{Objective}  & Data & \multicolumn{3}{c}{Classification models}  & \multicolumn{3}{c}{Regression models} & Baseline \\
& Collection & GP-C & SVM-C  & MLP-C   & GP-R & SVM-R  & MLP-R & BranGP\\ \cline{1-1} \cline{2-2} \cline{3-5} \cline{6-8} \cline{9-9} 
Cmax & 0.17   & 45.66  & 0.02   & 0.55 & 30.93 & 0.02 & 0.12    & 840.00 \\ 
Tmax & 0.21   & 47.21  & 0.02   & 0.38 & 28.65 & 0.02 & 0.10    & 300.00 \\
TWT  & 34.50  & 47.55  & 0.02   & 1.22 & 28.31 & 0.02 & 0.07    & 540.00 \\\bottomrule
\end{tabular}
\end{table*}

\begin{table}[!t]
\centering
\caption{The 5-fold cross validation accuracy for the classification models and the coefficient of determination ($R^2$) for the regression models. The best results are in bold.}
\label{tab:training_accuracy}
% \renewcommand{\arraystretch}{1.0}
% \resizebox{0.48\textwidth}{!}{
\begin{tabular}{@{\extracolsep{6pt}}lrrrrrr@{}}
\toprule
\multirow{2}{*}{Obj} & \multicolumn{3}{c}{Classification models}  & \multicolumn{3}{c}{Regression models} \\
   &  GP-C  & SVM-C    & MLP-C        & GP-R & SVM-R & MLP-R \\\cline{1-1} \cline{2-4}\cline{5-7}
Cmax  & 0.868  & 0.885  & \bf{0.890} &  0.845    & 0.843  & \bf{0.860} \\ 
Tmax  & 0.919  & \bf{0.924}  & 0.922      &  0.919    & 0.923  & \bf{0.925} \\
TWT   & 0.905 & 0.906  & \bf{0.911} & 0.856      & 0.876  & \bf{0.886}   \\\bottomrule
\end{tabular}
% }
\end{table}

\begin{table*}[!t]
\centering
\caption{The coefficients of the features in the decision functions learned by SVM-R and SVM-C. The coefficients with the top two largest magnitude are highlighted in bold.}
\label{tab::feature_coefficient}
\resizebox{\textwidth}{!}{
\begin{tabular}{llrrrrrrrrrrr}
\toprule
Obj & Method & $f_1$ (NPO) & $f_2$ (PT) & $f_3$ (PTB)  & $f_4$ (PTA) & $f_5$ (TPT) & $f_6$ (DD) & $f_7$ (W) & $f_8$ (RT) & $f_9$ (EST) & $f_{10}$ (WL) & bias  \\\midrule

\multirow{2}{*}{Cmax} & SVM-R & \bf{0.520} & -0.020 & 0.172 & -0.331 & -0.159 & 0.254 & -0.009 & 0.217 & \bf{0.390} & 0.003 & 0.159 \\
& SVM-C  & 1.026 & \bf{1.437} & 0.670 & \bf{-1.539} & -0.870 & 0.906 & -0.010 & 0.225 & 0.894 & 0.000 & -4e-10\\\midrule 

\multirow{2}{*}{Tmax} & SVM-R  & 0.026 & 0.224 & \bf{0.422} & -0.408 & 0.014 & 0.345 & -0.008 & 0.207 & \bf{0.629} & 0.007 & -0.080\\
& SVM-C  & 0.123 & \bf{2.846} & 1.481 & -1.562 & -0.081 & 1.572 & -0.103 & 0.880 & \bf{2.361} & 0.000 & -2e-08\\\midrule 

\multirow{2}{*}{TWT} & SVM-R & 0.011 & 0.217 & 0.370 & -0.427 & -0.057 & \bf{0.453} & -0.091 & 0.163 & \bf{0.533} & 0.016 & -0.057 \\
& SVM-C  & 0.159 & \bf{2.299} & 1.103 & -1.614 & -0.511 & \bf{2.192} & -0.727 & 0.292 & 1.395 & 0.000 & 2e-10 \\\bottomrule
\end{tabular}
}
\end{table*}

\subsection{Results of training}\label{subsec::training}

\textbf{Training set.} We use 100 randomly generated JSS problem instances of size $9\times9$ to construct our training sets. \added[]{The reason for choosing $9\times9$ instances is that the size is not too small, and the instances can be solved efficiently.} For each objective, Cmax, Tmax and TWT, we solve the training problem instances to optimally using OR-Tools. As shown in Table~\ref{tab::training time} (the column `Data Collection'), the training problem instances can be solved efficiently, especially for Cmax and Tmax. We construct a separate training set to train the regression and classification models for each of the objectives.

\textbf{Training time.} The time taken to train each ML model is shown in Table~\ref{tab::training time}. The linear model, SVM, can be trained in a few seconds, and MLP in about one minute. Training our GP model is slower but still can be done in less than one hour. Overall, the training of our supervised learning models is much more efficient than that of the evolutionary learning approach, BranGP.

\textbf{Training accuracy.} The 5-fold cross validation accuracy of the classification models and coefficient of determination ($R^2$) of the regression models are shown in Table~\ref{tab:training_accuracy}. All the ML algorithms achieve a reasonable accuracy and $R^2$ score for the classification and regression tasks. Overall, MLP performs the best and GP is slightly worse than the other two. BranGP is not a supervised learning model, so no classification accuracy or $R^2$ score can be reported.

\textbf{Model interpretability of SVM.} The coefficients of the features in the decision functions learned by SVM-C (for classification) and SVM-R (for regression) are shown in Table~\ref{tab::feature_coefficient}. Here, the magnitude of a coefficient indicates the importance of the corresponding feature, and the sign of a coefficient represents the correlation between the feature with the label. For example in the decision functions learned by SVM-C for Tmax, the features $f_2$ (PT) and $f_9$ (EST) are of the most importance as their coefficients have the largest magnitude. In addition, these two coefficients are positive, meaning that an operation with smaller processing time (PT) and smaller earliest start time (EST) should be scheduled earlier. Overall, the features $f_2$ (PT) and $f_9$ (EST) are two most important features for all three objectives. 

% \note[AE]{It seems a bit problematic that DD is included with a non-trivial weight for Cmax given that the due date cannot have any impact on the optimal solution here. Though for some of the data sets that don't even have a due date perhaps this doesn't have any effect anyway?}

\textbf{Model interpretability of GP.} The decision functions learned by GP-C (for classification) are 
\begin{itemize}
    \item Cmax: $f_1$;
    \item Tmax: $0.572f_2 + f_9$;
    \item TWT: $0.065 f_2 - f_7 + 0.090 f_9 - 0.004$.
\end{itemize}
The decision functions learned by GP-R (for regression) are: 
\begin{itemize}
    \item Cmax: $f_3 + f_9 / (f_3 + f_5 + 0.176)$;
    \item Tmax: $0.237 f_3 + 1.203 f_9$;
    \item TWT:  $1.271 f_9$.   
\end{itemize}
The acronyms of the features can be found in Table~\ref{tab::feature_coefficient}. It is surprising that most of the decision functions learned by GP are linear. For example, the one learned by GP-C for Cmax only uses one feature $f_1$ (number of precedent operations) to order variables. The earliest start time ($f_9$) is the most important feature used by five of six decision functions.

\begin{figure*}[!t]
    \centering
    \resizebox{\textwidth}{!}{
    \subfloat[Cmax]{
    \includegraphics[width=0.33\textwidth]{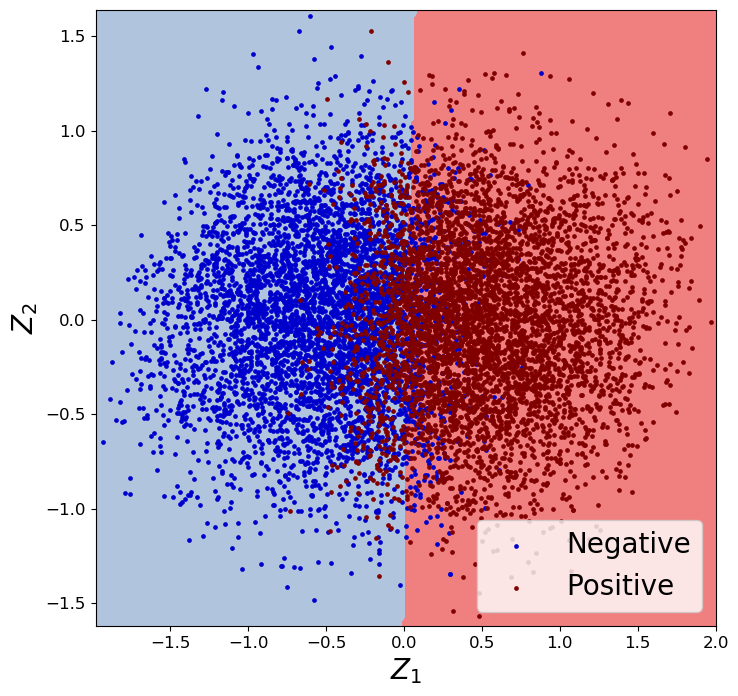}
    }
    \subfloat[Tmax]{
    \includegraphics[width=0.33\textwidth]{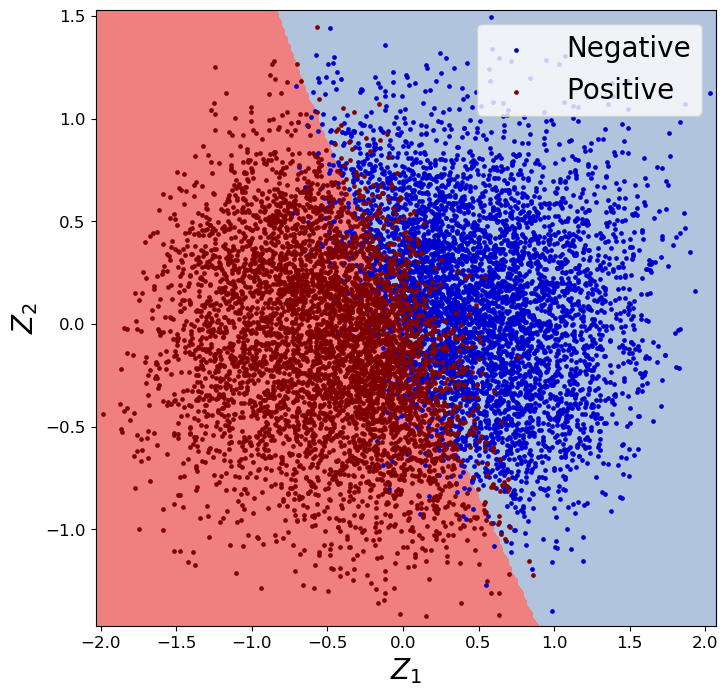}
    }
    \subfloat[TWT]{
    \includegraphics[width=0.33\textwidth]{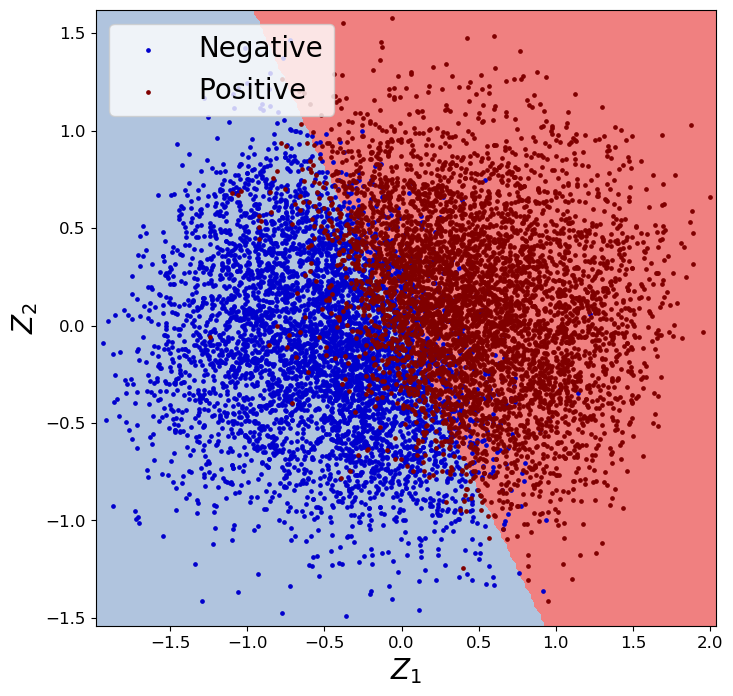}
    }
    }
    \caption{The decision boundary learned by MLP in the 2-D space spanned by the first two principal components ($Z_1$ and $Z_2$) of PCA analysis of the training data.}
    \label{fig:MLPboundary}
\end{figure*}

\textbf{Model interpretability of MLP.} MLP is a `black-box' neural network model. However we can visualize the decision boundary learned by MLP-C empirically. To do so, we use principal component analysis (PCA) to project the 10-D feature vectors onto a 2-D space, spanned by the first two principal components. MLP is then trained in the 2-D space, and the decision boundary learned is shown in Fig.~\ref{fig:MLPboundary}. These decision boundaries are also close to linear.

\begin{table*}[!t]
\centering
\caption{The number of branches required by each variable ordering method to solve the test problem instances to optimality. Each problem size has 100 instances. The best (smallest) mean values are highlighted in bold. The p-values are generated from the paired t-tests by comparing each method against the one with the best mean values (excluding OptSol). The p-values with statistical significance ($<0.05$) are in italic.}
\label{tab::MLbranches}
\resizebox{\textwidth}{!}{
\begin{tabular}{lllrrrrrrrrrrs}
\toprule
Obj                  & Size                          & Stats   & Default         & MinDom          & LowMin          & BranGP        & GP-C            & SVM-C           & MLP-C           & GP-R            & SVM-R           & MLP-R           & OptSol          \\\midrule
\multirow{15}{*}{Cmax} & \multirow{3}{*}{$6\times6$}   & mean    & 4.71e+02        & 3.17e+02        & 3.16e+02        & 2.10e+02      & 2.25e+02        & 2.35e+02        & 2.22e+02        & 2.32e+02        & 2.16e+02        & \bf{1.97e+02}   & 3.92e+01        \\
               &                         & std     & 2.94e+02        & 2.04e+02        & 2.24e+02        & 1.69e+02      & 1.66e+02        & 2.15e+02        & 2.01e+02        & 1.76e+02        & 1.85e+02        & 1.35e+02        & 1.02e+01        \\
               &                         & p-value & \emph{3.59e-16} & \emph{2.92e-07} & \emph{1.57e-06} & 4.33e-01      & 1.40e-01        & 8.19e-02        & 2.47e-01        & \emph{1.93e-02} & 1.66e-01        & -      & \emph{2.76e-20} \\\cline{2-14}
               & \multirow{3}{*}{$8\times8$}   & mean    & 2.14e+03        & 1.36e+03        & 1.21e+03        & \bf{7.59e+02} & 8.53e+02        & 8.79e+02        & 8.64e+02        & 1.02e+03        & 1.03e+03        & 8.13e+02        & 7.69e+01        \\
               &                         & std     & 1.19e+03        & 9.05e+02        & 8.35e+02        & 5.87e+02      & 6.57e+02        & 6.49e+02        & 5.55e+02        & 6.48e+02        & 6.86e+02        & 5.72e+02        & 3.70e+01        \\
               &                         & p-value & \emph{6.17e-20} & \emph{4.01e-09} & \emph{4.66e-06} & -    & 1.92e-01        & 1.11e-01        & 1.41e-01        & \emph{1.91e-04} & \emph{1.43e-03} & 4.36e-01        & \emph{3.16e-20} \\\cline{2-14}
               & \multirow{3}{*}{$10\times10$} & mean    & 6.60e+03        & 4.74e+03        & 4.54e+03        & \bf{2.55e+03} & 3.04e+03        & 3.44e+03        & 3.23e+03        & 3.65e+03        & 3.45e+03        & 3.12e+03        & 4.18e+02        \\
               &                         & std     & 2.75e+03        & 2.64e+03        & 2.37e+03        & 1.79e+03      & 1.80e+03        & 1.76e+03        & 1.72e+03        & 1.63e+03        & 1.72e+03        & 1.69e+03        & 8.90e+02        \\
               &                         & p-value & \emph{1.09e-24} & \emph{1.19e-13} & \emph{1.55e-10} & -    & \emph{9.04e-03} & \emph{5.13e-05} & \emph{1.84e-03} & \emph{1.56e-07} & \emph{5.59e-05} & \emph{9.82e-03} & \emph{1.34e-19} \\\cline{2-14}
               & \multirow{3}{*}{$12\times12$} & mean    & 1.61e+04        & 1.29e+04        & 1.20e+04        & 8.12e+03      & 8.30e+03        & 8.76e+03        & 8.90e+03        & 8.75e+03        & 9.02e+03        & \bf{7.58e+03}   & 5.96e+03        \\
               &                         & std     & 6.58e+03        & 8.09e+03        & 6.52e+03        & 5.98e+03      & 6.20e+03        & 7.07e+03        & 7.85e+03        & 6.40e+03        & 5.08e+03        & 6.21e+03        & 1.75e+04        \\
               &                         & p-value & \emph{3.23e-25} & \emph{1.50e-15} & \emph{1.05e-12} & 2.11e-01      & 7.48e-02        & \emph{8.31e-04} & \emph{5.95e-04} & \emph{1.05e-04} & \emph{1.89e-05} & -      & 2.23e-01        \\\cline{2-14}
               & \multirow{3}{*}{$14\times14$} & mean    & 3.71e+04        & 3.12e+04        & 2.68e+04        & 1.98e+04      & 2.08e+04        & 2.21e+04        & 2.16e+04        & 2.14e+04        & 2.12e+04        & \bf{1.92e+04}   & 4.88e+04        \\
               &                         & std     & 1.94e+04        & 2.30e+04        & 1.56e+04        & 1.64e+04      & 1.87e+04        & 2.06e+04        & 1.87e+04        & 1.59e+04        & 1.40e+04        & 1.81e+04        & 1.98e+05        \\
               &                         & p-value & \emph{7.52e-27} & \emph{1.76e-21} & \emph{8.25e-11} & 4.00e-01      & \emph{1.19e-02} & \emph{2.69e-04} & \emph{3.41e-02} & \emph{1.62e-04} & 5.64e-02        & -      & 1.20e-01  \\\midrule
   \multirow{15}{*}{Tmax} & \multirow{3}{*}{$6\times6$}   & mean    & 5.61e+02        & 6.84e+02        & 3.97e+02        & 2.73e+02        & 2.61e+02        & \bf{2.19e+02} & 2.23e+02 & 2.66e+02        & 2.53e+02        & 2.45e+02      & 4.62e+01        \\
                 &                         & std     & 4.02e+02        & 5.11e+02        & 3.60e+02        & 2.84e+02        & 1.83e+02        & 1.46e+02      & 1.45e+02 & 1.91e+02        & 1.98e+02        & 1.75e+02      & 1.55e+01        \\
                 &                         & p-value & \emph{4.11e-13} & \emph{6.62e-15} & \emph{1.13e-05} & 8.48e-02        & \emph{1.63e-02} & -    & 5.53e-01 & \emph{3.43e-02} & 1.29e-01        & 2.06e-01      & \emph{5.22e-21} \\\cline{2-14}
                 & \multirow{3}{*}{$8\times8$}   & mean    & 2.39e+03        & 2.76e+03        & 1.71e+03        & 1.13e+03        & 9.61e+02        & 9.95e+02      & 9.18e+02 & 9.60e+02        & \bf{9.03e+02}   & 1.00e+03      & 1.13e+02        \\
                 &                         & std     & 1.13e+03        & 1.33e+03        & 1.18e+03        & 7.74e+02        & 6.13e+02        & 7.30e+02      & 7.18e+02 & 7.00e+02        & 6.88e+02        & 8.54e+02      & 1.21e+02        \\
                 &                         & p-value & \emph{2.69e-24} & \emph{8.36e-24} & \emph{4.62e-09} & \emph{8.47e-03} & 4.70e-01        & 2.22e-01      & 8.40e-01 & 4.01e-01        & -      & 1.90e-01      & \emph{9.41e-21} \\\cline{2-14}
                 & \multirow{3}{*}{$10\times10$} & mean    & 8.12e+03        & 8.77e+03        & 5.12e+03        & 4.19e+03        & 3.10e+03        & 2.98e+03      & 3.10e+03 & 3.09e+03        & 3.07e+03        & \bf{2.77e+03} & 4.50e+02        \\
                 &                         & std     & 3.19e+03        & 3.67e+03        & 2.40e+03        & 2.24e+03        & 1.84e+03        & 1.78e+03      & 1.72e+03 & 1.71e+03        & 1.79e+03        & 1.56e+03      & 6.15e+02        \\
                 &                         & p-value & \emph{2.26e-29} & \emph{1.30e-29} & \emph{1.49e-15} & \emph{2.25e-08} & 8.79e-02        & 2.57e-01      & 6.92e-02 & \emph{2.76e-02} & \emph{3.40e-02} & -    & \emph{1.98e-26} \\\cline{2-14}
                 & \multirow{3}{*}{$12\times12$} & mean    & 1.87e+04        & 2.18e+04        & 1.35e+04        & 1.05e+04        & 7.79e+03        & 7.83e+03      & 7.91e+03 & 8.01e+03        & 8.14e+03        & \bf{7.70e+03} & 5.02e+03        \\
                 &                         & std     & 6.20e+03        & 8.01e+03        & 5.94e+03        & 4.72e+03        & 3.67e+03        & 3.83e+03      & 3.72e+03 & 4.00e+03        & 3.94e+03        & 3.66e+03      & 1.05e+04        \\
                 &                         & p-value & \emph{6.04e-32} & \emph{1.24e-35} & \emph{1.32e-15} & \emph{1.25e-09} & 8.07e-01        & 7.43e-01      & 5.15e-01 & 2.84e-01        & 1.01e-01        & -    & \emph{4.74e-03} \\\cline{2-14}
                 & \multirow{3}{*}{$14\times14$} & mean    & 4.91e+04        & 5.20e+04        & 3.75e+04        & 3.05e+04        & 2.92e+04        & 2.81e+04      & 2.79e+04 & 2.80e+04        & \bf{2.66e+04}   & 2.67e+04      & 8.72e+04        \\
                 &                         & std     & 2.18e+04        & 2.31e+04        & 2.69e+04        & 1.96e+04        & 2.36e+04        & 2.57e+04      & 2.52e+04 & 2.93e+04        & 2.17e+04        & 2.39e+04      & 1.48e+05        \\
                 &                         & p-value & \emph{2.08e-29} & \emph{3.17e-33} & \emph{5.77e-11} & \emph{1.27e-03} & \emph{3.18e-02} & 2.52e-01      & 3.20e-01 & 3.69e-01        & -      & 9.18e-01      & \emph{1.41e-05} \\\midrule
\multirow{15}{*}{TWT} & \multirow{3}{*}{$6\times6$}   & mean    & 1.11e+03        & 1.10e+03        & 8.35e+02      & 1.05e+03        & 8.49e+02        & 8.37e+02        & 8.18e+02        & \bf{7.83e+02}   & 8.04e+02 & 8.56e+02        & 5.38e+02        \\
                &                         & std     & 1.31e+03        & 1.25e+03        & 1.26e+03      & 1.67e+03        & 1.11e+03        & 1.03e+03        & 1.06e+03        & 9.60e+02        & 1.04e+03 & 1.27e+03        & 1.18e+03        \\
                &                         & p-value & \emph{4.95e-05} & \emph{4.60e-04} & 4.26e-01      & \emph{9.45e-03} & 3.25e-01        & 4.08e-01        & 5.77e-01        & -      & 6.95e-01 & 2.52e-01        & \emph{3.88e-05} \\\cline{2-14}
                & \multirow{3}{*}{$7\times7$}   & mean    & 3.23e+03        & 3.27e+03        & 2.36e+03      & 2.95e+03        & \bf{2.30e+03}   & 2.75e+03        & 2.38e+03        & 2.34e+03        & 2.35e+03 & 2.42e+03        & 2.37e+03        \\
                &                         & std     & 3.84e+03        & 2.83e+03        & 2.32e+03      & 3.80e+03        & 2.54e+03        & 3.24e+03        & 2.78e+03        & 2.76e+03        & 2.74e+03 & 2.74e+03        & 6.70e+03        \\
                &                         & p-value & \emph{2.00e-04} & \emph{1.30e-05} & 7.08e-01      & \emph{4.19e-03} & -      & \emph{1.42e-02} & 6.89e-01        & 8.33e-01        & 7.90e-01 & 3.51e-01        & 8.97e-01        \\\cline{2-14}
                & \multirow{3}{*}{$8\times8$}   & mean    & 1.56e+04        & 1.40e+04        & \bf{1.20e+04} & 1.64e+04        & 1.42e+04        & 1.35e+04        & 1.38e+04        & 1.35e+04        & 1.21e+04 & 1.27e+04        & 1.62e+04        \\
                &                         & std     & 1.79e+04        & 1.40e+04        & 1.36e+04      & 1.84e+04        & 1.67e+04        & 1.57e+04        & 1.99e+04        & 1.66e+04        & 1.36e+04 & 1.65e+04        & 2.43e+04        \\
                &                         & p-value & \emph{2.69e-05} & \emph{2.84e-02} & -    & \emph{7.03e-05} & \emph{1.19e-02} & 1.20e-01        & 8.03e-02        & \emph{3.58e-02} & 8.59e-01 & 3.70e-01        & \emph{4.18e-03} \\\cline{2-14}
                & \multirow{3}{*}{$9\times9$}   & mean    & 8.50e+04        & 7.11e+04        & 6.70e+04      & 1.01e+05        & 7.83e+04        & 7.82e+04        & 6.57e+04        & 6.88e+04        & 6.88e+04 & \bf{6.44e+04}   & 1.25e+05        \\
                &                         & std     & 1.28e+05        & 9.48e+04        & 1.16e+05      & 1.54e+05        & 1.62e+05        & 1.29e+05        & 9.64e+04        & 1.13e+05        & 1.10e+05 & 1.01e+05        & 1.81e+05        \\
                &                         & p-value & \emph{5.64e-03} & 2.08e-01        & 6.27e-01      & \emph{1.60e-05} & 1.93e-01        & \emph{4.76e-02} & 8.02e-01        & 2.13e-01        & 2.28e-01 & -      & \emph{1.58e-07} \\\cline{2-14}
                & \multirow{3}{*}{$10\times10$} & mean    & 4.14e+05        & 3.69e+05        & \bf{3.19e+05} & 4.58e+05        & 3.65e+05        & 4.23e+05        & 3.75e+05        & 3.78e+05        & 3.54e+05 & 4.05e+05        & 5.84e+05        \\
                &                         & std     & 4.97e+05        & 4.74e+05        & 3.96e+05      & 4.89e+05        & 4.39e+05        & 5.48e+05        & 4.26e+05        & 4.79e+05        & 4.33e+05 & 5.87e+05        & 6.77e+05        \\
                &                         & p-value & \emph{3.54e-04} & \emph{2.71e-02} & -    & \emph{2.48e-07} & \emph{3.07e-02} & \emph{7.67e-05} & \emph{1.60e-02} & \emph{1.24e-02} & 1.38e-01 & \emph{3.52e-03} & \emph{4.19e-09} \\\bottomrule
\end{tabular}
}
\end{table*}

\subsection{Efficacy of ML-based variable ordering methods}

We evaluate the efficacy of our ML-based variable ordering methods on randomly generated JSS problem instances. For Cmax and Tmax, we generate 100 instances for each problem size in \{$6\times6$, $8\times8$, $10\times10$, $12\times12$, $14\times14$\}. For TWT, the test problem sizes are \{$6\times6$, $7\times7$, $8\times8$, $9\times9$, $10\times10$\} because the TWT objective is much hard than the other two to solve. We use the CP-SAT solver in OR-Tools with ten different variable ordering methods to solve the test problem instances. As the test instances can be optimally solved, we also investigate a variable ordering strategy, called \emph{OptSol}, that orders variables based on the optimal solution values. 

The mean and standard deviation of the number of branches used by each method to optimally solve the problem instances are presented in Table~\ref{tab::MLbranches}. Here, we also present the p-values of the paired t-tests comparing each method against the one with the best mean values. Overall, our methods (GP-C, SVM-C, MLP-C, GP-R, SVM-R, and MLP-R) generate the best results compared to the four baselines (Default, MinDom, LowMin and BranGP). Specifically, all our six classification and regression models significantly outperforms the Default and MinDom methods for all the three objectives. The LowMin method performs very well for TWT, but it is significantly outperformed by our methods for the other two objectives. For some of the instances, including those TWT instances where LowMin performs well, the OptSol approach does not perform very well. This demonstrates the limitation of using the optimal solutions from small instances as training data. While ordering variables based on an optimal solution guarantees that the optimal solution is found at the start of the search and hence that the best possible upper bound is available, it does not guarantee that the size of the search tree is minimised. Similarly, the BranGP method is very competitive for Cmax, but it is consistently outperformed by our methods for Tmax and TWT.

\begin{figure*}[!t]
    \centering
    \resizebox{\textwidth}{!}{
    \subfloat[Cmax]{
    \includegraphics[width=0.33\textwidth]{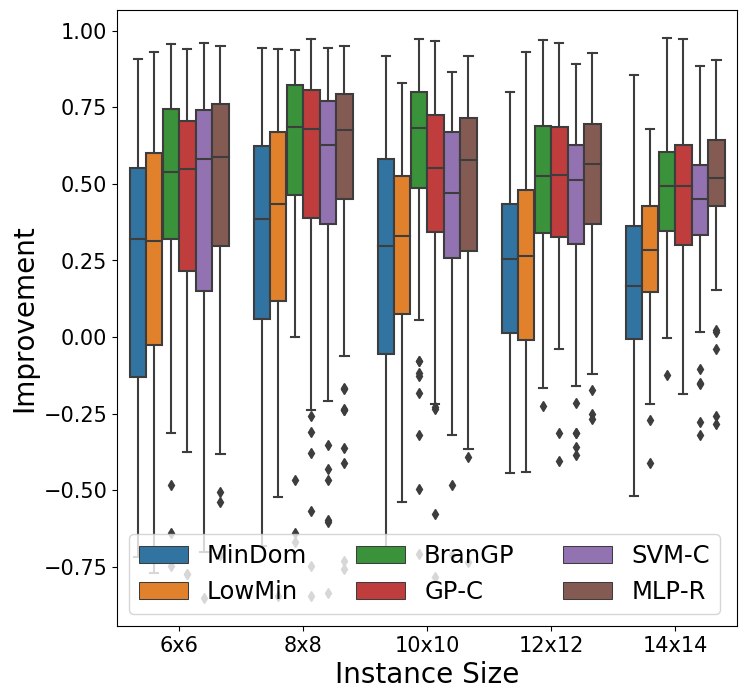}
    }
    \subfloat[Tmax]{
    \includegraphics[width=0.33\textwidth]{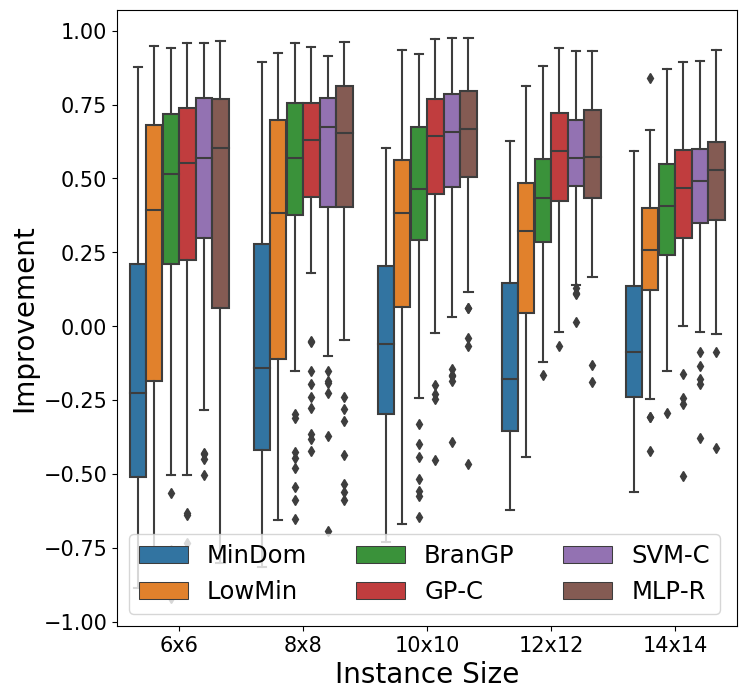}
    }
    \subfloat[TWT]{
    \includegraphics[width=0.33\textwidth]{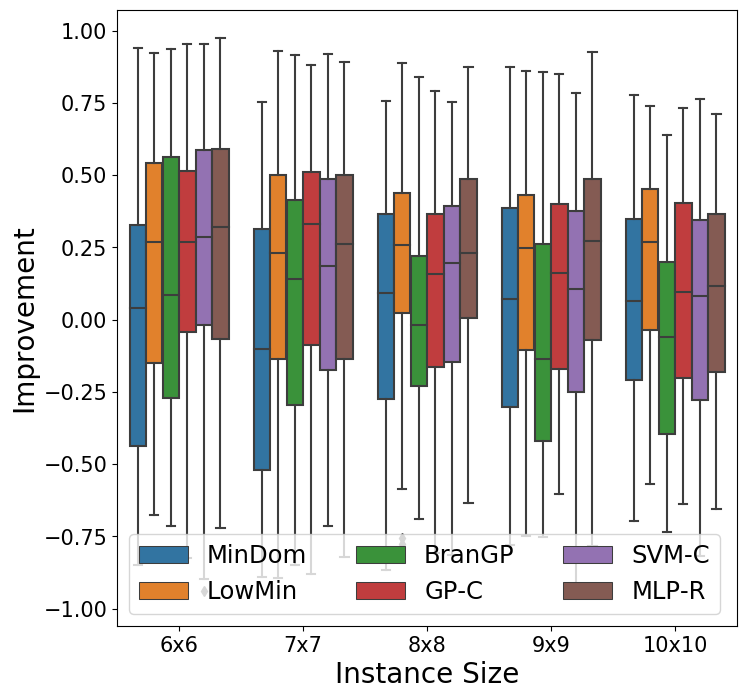}
    }
    }
    \caption{The comparison between our selected variable ordering methods (GP-C, SVM-C, MLP-R) and baselines, in terms of the number of branches required to solve the test instances. The improvement is computed over the Default method.}
    \label{fig:imp_small_branch}
\end{figure*}
\begin{figure*}[!t]
    \centering
    \resizebox{\textwidth}{!}{
    \subfloat[Cmax]{
    \includegraphics[width=0.33\textwidth]{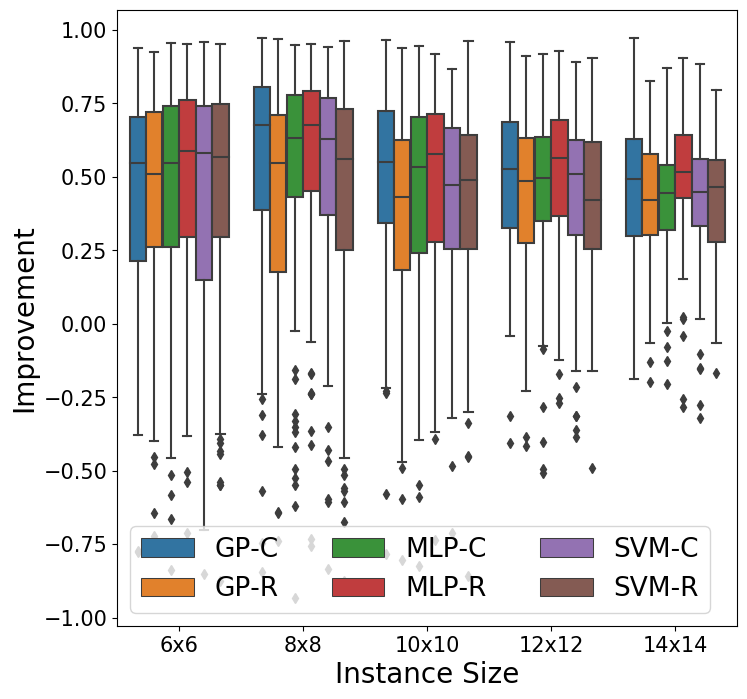}
    }
    \subfloat[Tmax]{
    \includegraphics[width=0.33\textwidth]{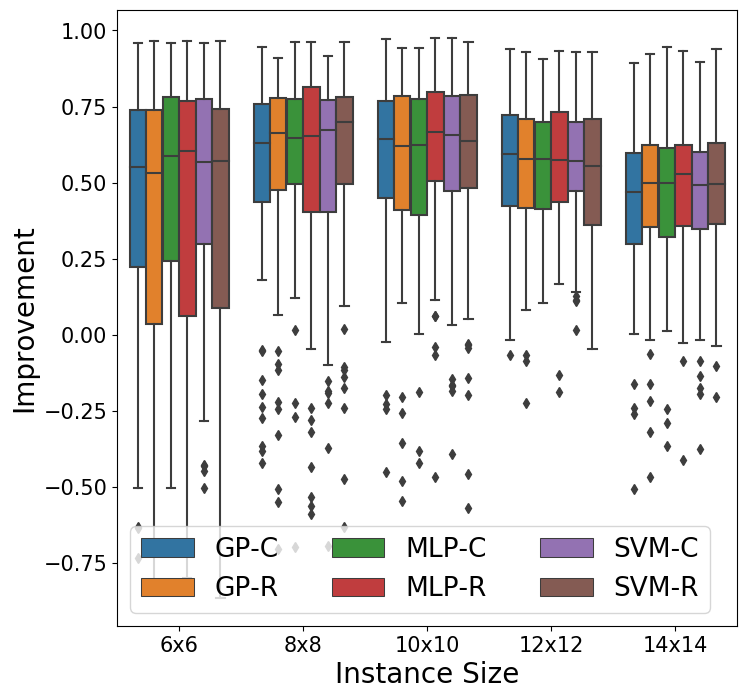}
    }
    \subfloat[TWT]{
    \includegraphics[width=0.33\textwidth]{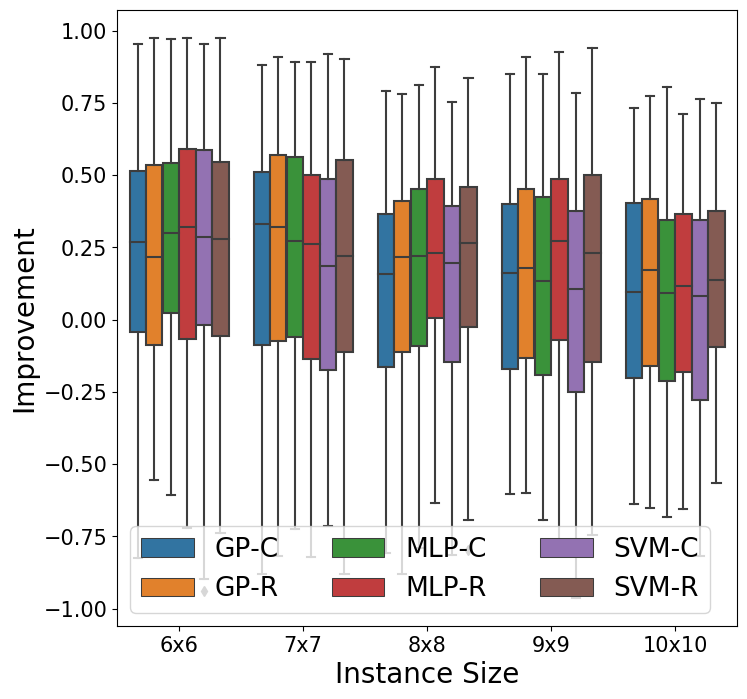}
    }
    }
    \caption{The comparison between our classification and regression based methods, in terms of the number of branches required to solve the test instances. The improvement is computed over the Default method.}
    \label{fig:ml_models}
\end{figure*}

\begin{figure*}[!t]
    \centering
    \resizebox{\textwidth}{!}{
    \subfloat[Cmax]{
    \includegraphics[width=0.33\textwidth]{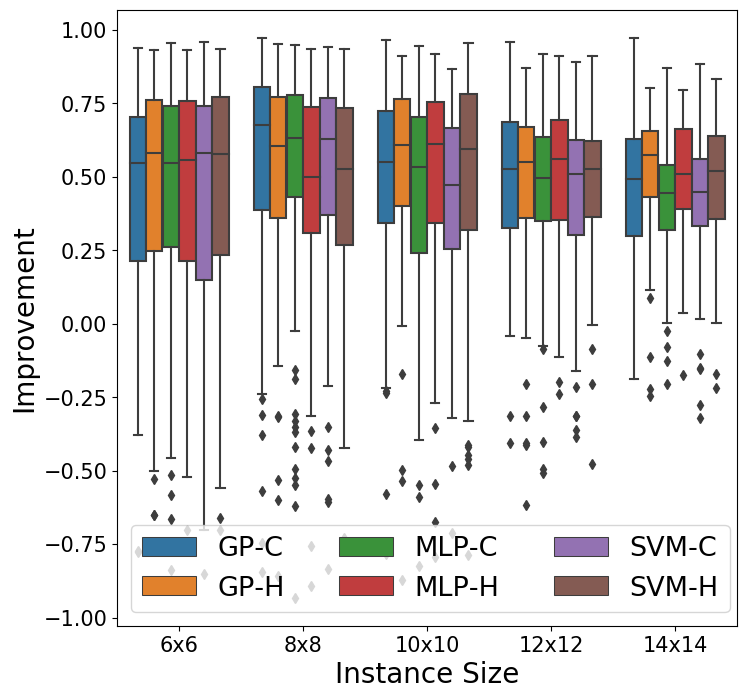}
    }
    \subfloat[Tmax]{
    \includegraphics[width=0.33\textwidth]{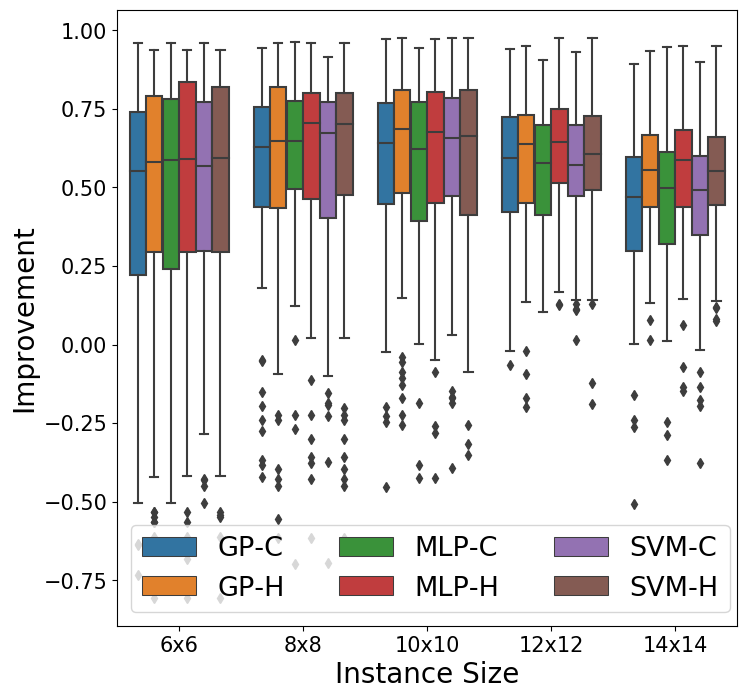}
    }
    \subfloat[TWT]{
    \includegraphics[width=0.33\textwidth]{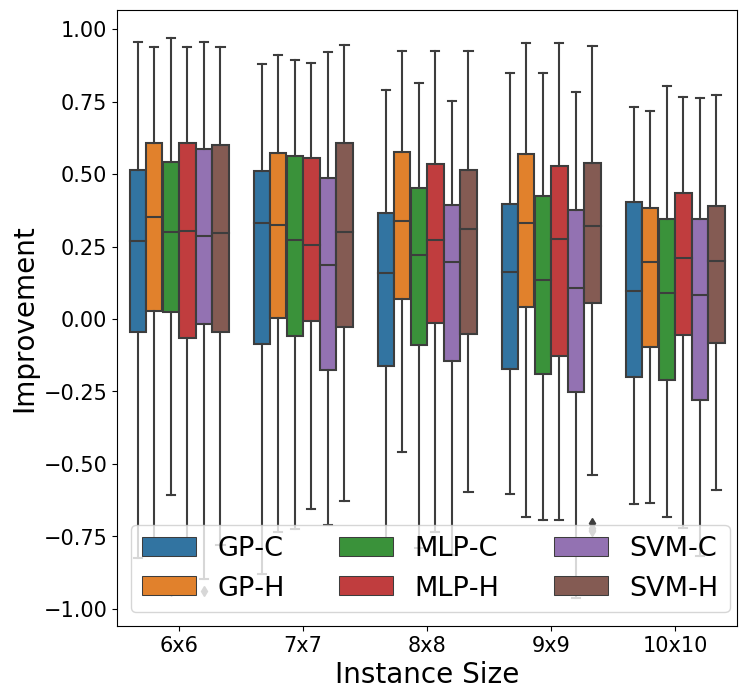}
    }
    }
    \caption{The comparison between pure ML-based (-C) and hybrid (-H) variable ordering methods, in terms of the number of branches required to solve the test instances. The improvement is computed over the Default method.}
    \label{fig:hybrid}
\end{figure*}

In Fig.~\ref{fig:imp_small_branch}, we select three of our methods (GP-C, SVM-C, MLP-R) and compare them to the baselines using box plots. Here, for each variable ordering method, we compute an improvement over the Default method as $(n_d - n) / \max \{ n_d,  n \}$, where $n_d$ and $n$ are the number of branches used by the Default and the method under comparison respectively. An improvement value greater than zero indicates that the method performs better than the Default. We can observe that our ML-based methods generally produce the largest improvement compared to the baselines with only a few exceptions.

% \note[AE]{Looking at this again I wonder whether the problem with TWT is that we don't have a feature that measures weight relative to due date so something like $d_j/w_j$ or perhaps $w_j (M-d)$ for some suitable $M\approx C_{\max}$. No need to do anything here though.}

In Fig.~\ref{fig:ml_models}, we compare our six classification and regression models using box plots. In general, GP-C and MLP-R perform slightly better than others for Cmax, while all methods perform similarly for Tmax and TWT. This is also supported by the statistical test results presented in Table~\ref{tab::MLbranches}. Hence, our ML-based variable ordering method can significantly benefit the CP solver, no matter which of the learning algorithms is used.

The OptSol method, that orders variables based on the optimal solution, performs very well on small problem instances, but its performance deteriorates when the problem size increases. This may be because the OptSol method (and hence all our ML-based methods) focuses on finding high-quality solutions without taking into account how to efficiently prove optimality. The hybrid variable ordering methods should be able to alleviate this issue, because they combine the merits of both ML-based and domain-based variable ordering methods.

\begin{table*}[!t]
\centering
\caption{The number of branches used by the hybrid variable ordering methods to optimally solve the test problem instances. The best (smallest) mean values are in bold. The p-values are generated from the paired t-tests by comparing each method against the one with the best mean values (excluding OptSol). The p-values with statistical significance ($<0.05$) are in italic.}
\label{tab::hybrid_branch}
\resizebox{\textwidth}{!}{
\begin{tabular}{lllrrrrrrrs}
\toprule
Obj                    & Size                          & Stats   & Default         & MinDom          & LowMin          & BranGP-H      & GP-H        & SVM-H           & MLP-H         & OptSol-H        \\\midrule
\multirow{15}{*}{Cmax} & \multirow{3}{*}{$6\times6$}   & mean    & 4.71e+02        & 3.17e+02        & 3.16e+02        & 2.22e+02      & 2.08e+02        & \bf{2.07e+02}   & 2.18e+02      & 1.82e+02        \\
                 &                         & std     & 2.94e+02        & 2.04e+02        & 2.24e+02        & 1.71e+02      & 1.50e+02        & 1.59e+02        & 1.64e+02      & 1.44e+02        \\
                 &                         & p-value & \emph{2.54e-13} & \emph{1.36e-05} & \emph{1.84e-06} & 2.90e-01      & 9.44e-01        & -      & 2.09e-01      & 5.94e-02        \\\cline{2-11}
                 & \multirow{3}{*}{$8\times8$}   & mean    & 2.14e+03        & 1.36e+03        & 1.21e+03        & \bf{8.04e+02} & 8.47e+02        & 9.43e+02        & 9.19e+02      & 6.77e+02        \\
                 &                         & std     & 1.19e+03        & 9.05e+02        & 8.35e+02        & 5.54e+02      & 5.97e+02        & 5.87e+02        & 6.21e+02      & 4.89e+02        \\
                 &                         & p-value & \emph{7.66e-19} & \emph{1.90e-08} & \emph{4.52e-06} & -    & 4.88e-01        & \emph{3.62e-02} & 7.91e-02      & \emph{4.67e-02} \\\cline{2-11}
                 & \multirow{3}{*}{$10\times10$} & mean    & 6.60e+03        & 4.74e+03        & 4.54e+03        & 2.70e+03      & \bf{2.67e+03}   & 2.91e+03        & 2.85e+03      & 2.42e+03        \\
                 &                         & std     & 2.75e+03        & 2.64e+03        & 2.37e+03        & 1.43e+03      & 1.46e+03        & 1.65e+03        & 1.43e+03      & 1.55e+03        \\
                 &                         & p-value & \emph{5.40e-25} & \emph{4.25e-12} & \emph{8.23e-11} & 8.74e-01      & -      & 1.67e-01        & 2.46e-01      & 1.39e-01        \\\cline{2-11}
                 & \multirow{3}{*}{$12\times12$} & mean    & 1.61e+04        & 1.29e+04        & 1.20e+04        & \bf{7.08e+03} & 7.76e+03        & 7.87e+03        & 7.51e+03      & 6.49e+03        \\
                 &                         & std     & 6.58e+03        & 8.09e+03        & 6.52e+03        & 5.02e+03      & 4.66e+03        & 5.22e+03        & 4.44e+03      & 5.07e+03        \\
                 &                         & p-value & \emph{8.41e-32} & \emph{2.84e-18} & \emph{1.69e-15} & -    & \emph{3.19e-02} & \emph{3.98e-03} & 1.73e-01      & 8.58e-02        \\\cline{2-11}
                 & \multirow{3}{*}{$14\times14$} & mean    & 3.71e+04        & 3.12e+04        & 2.68e+04        & 1.90e+04      & 1.90e+04        & 1.85e+04        & \bf{1.84e+04} & 1.57e+04        \\
                 &                         & std     & 1.94e+04        & 2.30e+04        & 1.56e+04        & 1.90e+04      & 1.94e+04        & 1.23e+04        & 1.22e+04      & 1.27e+04        \\
                 &                         & p-value & \emph{6.94e-26} & \emph{1.99e-14} & \emph{1.37e-15} & 5.29e-01      & 5.16e-01        & 8.08e-01        & -    & \emph{1.26e-04} \\\midrule

\multirow{15}{*}{Tmax} & \multirow{3}{*}{$6\times6$}   & mean    & 5.61e+02        & 6.84e+02        & 3.97e+02        & 2.29e+02        & 2.37e+02      & \bf{2.28e+02}   & 2.28e+02      & 2.03e+02        \\
                 &                         & std     & 4.02e+02        & 5.11e+02        & 3.60e+02        & 1.92e+02        & 2.32e+02      & 2.23e+02        & 2.31e+02      & 2.18e+02        \\
                 &                         & p-value & \emph{1.85e-10} & \emph{3.93e-13} & \emph{1.16e-05} & 9.24e-01        & 4.23e-01      & -      & 9.64e-01      & 7.14e-02        \\\cline{2-11}
                 & \multirow{3}{*}{$8\times8$}   & mean    & 2.39e+03        & 2.76e+03        & 1.71e+03        & 1.03e+03        & 9.88e+02      & 9.59e+02        & \bf{9.53e+02} & 7.64e+02        \\
                 &                         & std     & 1.13e+03        & 1.33e+03        & 1.18e+03        & 7.87e+02        & 8.04e+02      & 8.21e+02        & 8.50e+02      & 6.20e+02        \\
                 &                         & p-value & \emph{2.41e-24} & \emph{4.51e-21} & \emph{5.99e-09} & 2.03e-01        & 5.52e-01      & 8.88e-01        & -    & \emph{1.98e-02} \\\cline{2-11}
                 & \multirow{3}{*}{$10\times10$} & mean    & 8.12e+03        & 8.77e+03        & 5.12e+03        & 3.33e+03        & \bf{2.73e+03} & 3.02e+03        & 2.93e+03      & 2.57e+03        \\
                 &                         & std     & 3.19e+03        & 3.67e+03        & 2.40e+03        & 1.80e+03        & 1.54e+03      & 1.89e+03        & 1.83e+03      & 1.67e+03        \\
                 &                         & p-value & \emph{3.55e-28} & \emph{1.78e-28} & \emph{7.15e-15} & \emph{7.59e-04} & -    & \emph{3.14e-02} & 1.68e-01      & 3.01e-01        \\\cline{2-11}
                 & \multirow{3}{*}{$12\times12$} & mean    & 1.87e+04        & 2.18e+04        & 1.35e+04        & 8.01e+03        & 7.29e+03      & 7.23e+03        & \bf{6.82e+03} & 6.73e+03        \\
                 &                         & std     & 6.20e+03        & 8.01e+03        & 5.94e+03        & 4.11e+03        & 3.75e+03      & 3.32e+03        & 3.58e+03      & 3.85e+03        \\
                 &                         & p-value & \emph{2.49e-33} & \emph{1.13e-34} & \emph{5.47e-22} & \emph{8.36e-04} & 1.67e-01      & 6.13e-02        & -    & 7.97e-01        \\\cline{2-11}
                 & \multirow{3}{*}{$14\times14$} & mean    & 4.91e+04        & 5.20e+04        & 3.75e+04        & 2.40e+04        & \bf{2.34e+04} & 2.41e+04        & 2.36e+04      & 2.35e+04        \\
                 &                         & std     & 2.18e+04        & 2.31e+04        & 2.69e+04        & 1.77e+04        & 1.61e+04      & 1.96e+04        & 2.04e+04      & 2.08e+04        \\
                 &                         & p-value & \emph{1.05e-35} & \emph{8.70e-33} & \emph{8.23e-14} & 4.36e-01        & -    & 4.52e-01        & 8.84e-01      & 9.36e-01  \\\midrule
\multirow{15}{*}{TWT} & \multirow{3}{*}{$6\times6$}   & mean    & 1.11e+03        & 1.10e+03        & 8.35e+02        & 7.47e+02      & \bf{6.97e+02} & 7.47e+02      & 7.64e+02 & 6.80e+02        \\
                &                         & std     & 1.31e+03        & 1.25e+03        & 1.26e+03        & 1.02e+03      & 8.57e+02      & 9.09e+02      & 9.61e+02 & 7.76e+02        \\
                &                         & p-value & \emph{1.59e-06} & \emph{1.24e-05} & 7.45e-02        & 4.36e-01      & -    & 1.46e-01      & 5.61e-02 & 7.18e-01        \\\cline{2-11}
                & \multirow{3}{*}{$7\times7$}   & mean    & 3.23e+03        & 3.27e+03        & 2.36e+03        & 2.13e+03      & \bf{1.92e+03} & 2.04e+03      & 2.20e+03 & 2.29e+03        \\
                &                         & std     & 3.84e+03        & 2.83e+03        & 2.32e+03        & 2.46e+03      & 2.02e+03      & 2.39e+03      & 2.72e+03 & 3.10e+03        \\
                &                         & p-value & \emph{4.46e-06} & \emph{1.25e-10} & \emph{4.74e-03} & 1.03e-01      & -    & 3.83e-01      & 5.60e-02 & \emph{2.86e-02} \\\cline{2-11}
                & \multirow{3}{*}{$8\times8$}   & mean    & 1.56e+04        & 1.40e+04        & 1.20e+04        & 1.14e+04      & 1.13e+04      & \bf{1.07e+04} & 1.23e+04 & 1.11e+04        \\
                &                         & std     & 1.79e+04        & 1.40e+04        & 1.36e+04        & 1.40e+04      & 1.55e+04      & 1.21e+04      & 1.78e+04 & 1.39e+04        \\
                &                         & p-value & \emph{4.71e-07} & \emph{7.66e-05} & \emph{4.80e-02} & 3.16e-01      & 4.43e-01      & -    & 9.40e-02 & 6.09e-01        \\\cline{2-11}
                & \multirow{3}{*}{$9\times9$}   & mean    & 8.50e+04        & 7.11e+04        & 6.70e+04        & 5.85e+04      & 5.90e+04      & \bf{5.64e+04} & 5.82e+04 & 6.15e+04        \\
                &                         & std     & 1.28e+05        & 9.48e+04        & 1.16e+05        & 1.08e+05      & 8.89e+04      & 9.98e+04      & 8.32e+04 & 9.50e+04        \\
                &                         & p-value & \emph{3.03e-03} & 5.08e-02        & \emph{4.55e-03} & 7.37e-01      & 6.55e-01      & -    & 7.36e-01 & 4.86e-01        \\\cline{2-11}
                & \multirow{3}{*}{$10\times10$} & mean    & 4.14e+05        & 3.69e+05        & 3.19e+05        & \bf{3.19e+05} & 3.45e+05      & 3.52e+05      & 3.33e+05 & 3.34e+05        \\
                &                         & std     & 4.97e+05        & 4.74e+05        & 3.96e+05        & 4.34e+05      & 4.33e+05      & 4.94e+05      & 4.33e+05 & 3.87e+05        \\
                &                         & p-value & \emph{1.72e-03} & 6.33e-02        & 9.76e-01        & -    & 2.44e-01      & 1.58e-01      & 4.56e-01 & 5.38e-01       \\\bottomrule
\end{tabular}
}
\end{table*}

\begin{figure*}[!t]
    \centering
    \resizebox{0.9\textwidth}{!}{
    \subfloat[Cmax]{
    \includegraphics[width=0.33\textwidth]{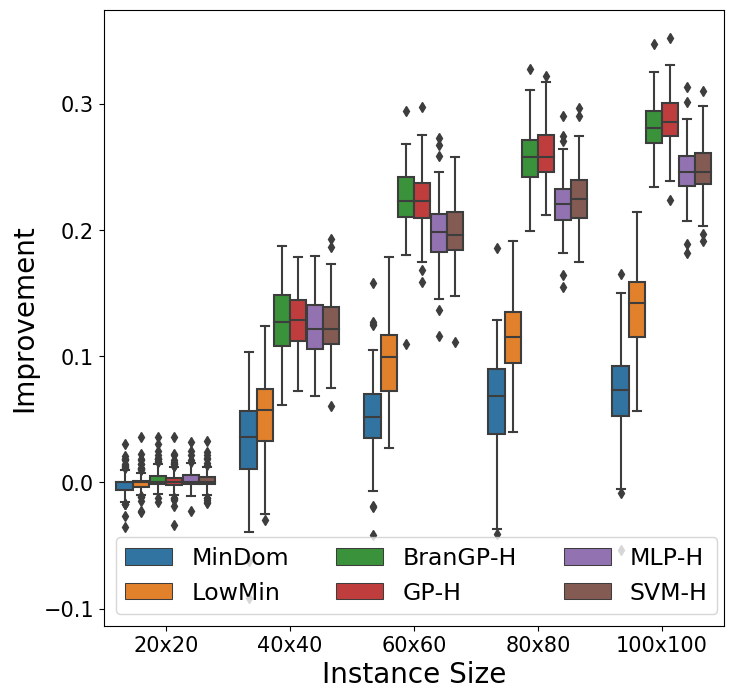}
    }
    \subfloat[Tmax]{
    \includegraphics[width=0.33\textwidth]{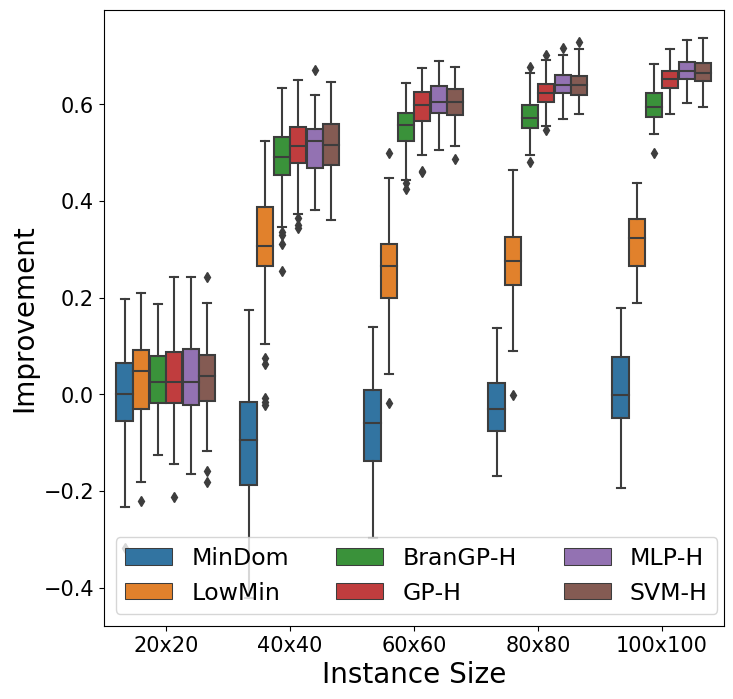}
    }
    \subfloat[TWT]{
    \includegraphics[width=0.33\textwidth]{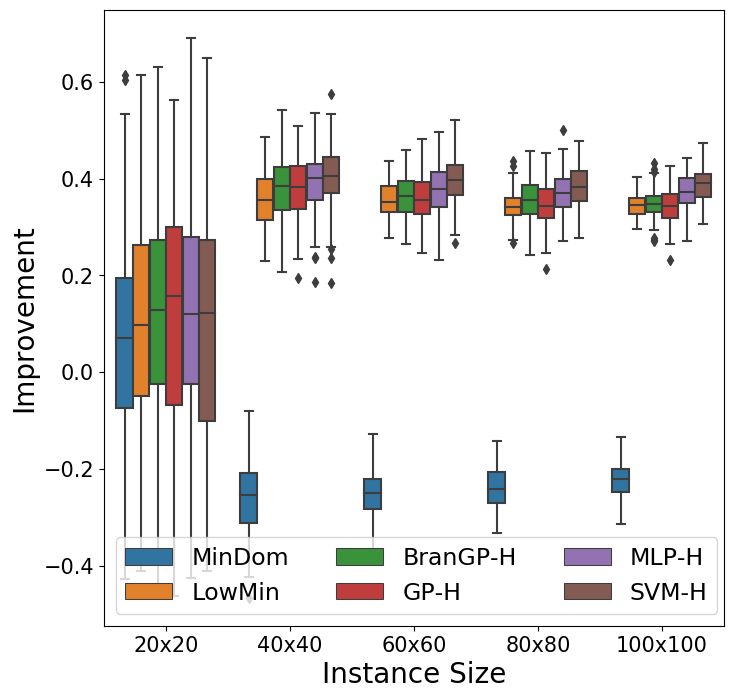}
    }
    }
    \caption{The comparison between our hybrid methods (GP-H, MLP-H, SVM-H) and baselines, in terms of the best objective value found within 60 seconds for larger problem instances. The improvement is computed over the Default method.}
    \label{fig:large_best}
\end{figure*}

\subsection{Efficacy of hybrid variable ordering methods}
\label{subsec::hybrid}

We hybridise our ML-based variable ordering methods with the domain-based LowMin method, because LowMin outperforms the MinDom and Default methods according to Table~\ref{tab::MLbranches}. Further, we will only consider the classification models (i.e., GP-C, SVM-C and MLP-C) from now on, since the regression models are not significantly different from the classification models. For a fair comparison, we also hybridise the baseline, BranGP, with LowMin. 

Fig.~\ref{fig:hybrid} shows the comparison between the pure ML-based methods (GP-C, SVM-C and MLP-C) against the hybrid variable ordering methods (GP-H, SVM-H and MLP-H). We can observe that the hybrid methods generally perform better than the corresponding ML-based methods. When the problem size increases, the difference between the hybrid and ML-based methods becomes more significant. This result is consistent with our intuition that the hybrid methods should be able to prove optimality faster especially on larger problem instances. 

Table~\ref{tab::hybrid_branch} presents the results of all hybrid methods and the baselines. Here, the hybrid methods consistently perform better than the Default, MinDom and LowMin across different problem sizes and objectives, in terms of the average number of branches used to solve the problem instances. The hybrid method using BranGP (BranGP-H) is competitive compared to our hybrid methods (GP-H, SVM-H and MLP-H).

\subsection{Generalisation to larger instances}
\label{subsec::testing_large}

We test the generalisation of our methods on larger problem instances. As the hybrid methods perform better than the pure ML-based methods, we will only test the hybrid methods in the rest of the paper. For each problem size in \{$20\times20$, $40\times40$, $60\times60$, $80\times80$, $100\times100$\}, we randomly generate 100 instances with different random seeds, similar as before. These instances are significantly larger, and they generally cannot be solved to optimality within a reasonable amount of time. Hence, we give 60 seconds to the CP-SAT solver and record the best objective values found within the cutoff time. 

Fig.~\ref{fig:large_best} shows the comparison between our hybrid methods (GP-H, MLP-H, SVM-H) and baselines. Here, the improvement is computed over the Default method in terms of the best objective value found. We can see that all methods perform better than the Default except MinDom. The LowMin method performs well for TWT, but is significantly outperformed by our hybrid methods. The BranGP-H method performs similarly with GP-H and better than MLP-H and SVM-H for Cmax. However, for the other two objectives Tmax and TWT, the BranGP-H method is generally outperformed by our hybrid methods, and the performance gap becomes larger when the problem size increases.

\begin{table*}[!t]
\centering
\caption{The comparison between our hybrid variable ordering methods (GP-H, SVM-H, MLP-H) against baselines on the benchmark instances for Cmax. The number of branches (Branch) are averaged across the instances that can be optimally solved by the Default method, and the best objective values found (objVal) are averaged across the remaining instances.}
\label{tab::benchmark_cmax}
\resizebox{\textwidth}{!}{
\begin{tabular}{llllllllll}
\toprule
Benchmark            & \#Instance          & Stats    & Default  & MinDom   & LowMin        & BranGP-H        & GP-H          & SVM-H    & MLP-H         \\\midrule
\multirow{3}{*}{ta}  & \multirow{3}{*}{80} & \#solved & 13       & 15       & 16            & 18            & \bf{20}            & 19       & 18            \\
               &               & branch   & 4.44e+05 & 3.69e+05 & 2.82e+05      & 2.65e+05      & 2.42e+05      & 2.56e+05 & \bf{2.34e+05} \\
               &               & objVal   & 2.79e+03 & 2.78e+03 & 2.67e+03      & \bf{2.61e+03}      & \bf{2.61e+03} & \bf{2.61e+03} & \bf{2.61e+03}      \\\midrule
\multirow{3}{*}{la}  & \multirow{3}{*}{40} & \#solved & 39       & 39       & 39            & 39            & 39            & 39       & 39            \\
               &               & Branch   & 7.60e+04 & 6.33e+04 & 4.25e+04      & 3.65e+04      & 3.18e+04      & 2.91e+04 & \bf{2.85e+04} \\
               &               & ObjVal   & \bf{1.15e+03} & 1.17e+03 & \bf{1.15e+03} & 1.16e+03      & 1.16e+03      & \bf{1.15e+03} & 1.16e+03      \\\midrule
\multirow{3}{*}{abz} & \multirow{3}{*}{5}  & \#solved & 2        & 2        & 2             & 2             & 2             & 2        & 2             \\
               &               & branch   & 1.32e+04 & 1.27e+04 & 7.59e+03      & \bf{5.44e+03} & 6.01e+03      & 6.52e+03 & 6.84e+03      \\
               &               & objVal   & 6.77e+02 & 6.86e+02 & 6.79e+02      & 6.77e+02      & 6.77e+02      & 6.76e+02 & \bf{6.74e+02} \\\midrule
\multirow{3}{*}{orb} & \multirow{3}{*}{10} & \#solved & 10       & 10       & 10            & 10            & 10            & 10       & 10            \\
               &               & branch   & 3.36e+04 & 2.54e+04 & 2.85e+04      & 2.31e+04      & 2.52e+04      & 2.14e+04 & \bf{1.88e+04} \\
               &               & objVal   & - & -      & -           & -           & -           & -      & -           \\\midrule
\multirow{3}{*}{swv} & \multirow{3}{*}{20} & \#solved & \bf{6}        & \bf{6}        & 5             & \bf{6}             & \bf{6}             & \bf{6}        & \bf{6}             \\
               &               & branch   & 7.51e+05 & 7.15e+05 & 2.05e+05      & 1.11e+05      & 8.20e+04      & 1.61e+05 & \bf{7.67e+04} \\
               &               & objVal   & \bf{2.18e+03} & 2.25e+03 & \bf{2.18e+03} & 2.19e+03      & 2.20e+03      & 2.21e+03 & 2.19e+03 \\\bottomrule    
\end{tabular}
}
\end{table*}

\subsection{Generalisation to benchmark instances}
\label{subsec::testing_benchmark}
We further test the generalisation of our hybrid variable ordering methods on benchmark instances that are very different from those used in training. The cutoff time for the CP-SAT solver to solve each problem instance is set to ten minutes. The results for the Cmax benchmarks are presented in Table~\ref{tab::benchmark_cmax}. Overall, our hybrid methods (GP-H, SVM-H and MLP-H) perform better than the Default, MinDow and LowMin methods. Our GP-H method performs the best in terms of the number of instances solved. Our MLP-H method generally performs the best in terms of the number of branches required for solved instances. All hybrid methods (including BranGP-H) perform similarly in terms of the best objective value found for unsolved instances. 

% The results for the TWT benchmark are presented in Table~\ref{tab::benchmark_twt}. Our hybrid methods perform well compared to the Default, MinDom and BranGP-H methods, and slightly worse than the LowMin method in terms of the number of branches used. Overall, the learning-based variable ordering methods do not have significant advantage over the traditional domain-based methods on these benchmark instances. This may be because the benchmark instances tested are very different from the training instances. This suggests that the ML models should be re-trained when the test instances are very different from the training instances.   

%% file: 06-conclusion.tex
\section{Conclusion}\label{sec:conclusion}

In this paper, we addressed the importance of variable ordering strategy in constraint programming (CP) and proposed a supervised learning approach to automatically learn effective variable ordering strategies. We used the job shop scheduling problems as a case study and aimed to order variables based on the predicted optimal solution of a problem instance. We demonstrated that solution prediction for job shop scheduling can be modeled as either a classification or regression task. We developed fast and accurate supervised learning models with three different algorithms, which outperformed an evolutionary learning approach. Our ML-based variable ordering methods were extensively evaluated against four baselines and showed better performance. Finally, we demonstrated the advantages of hybridizing our ML-based variable ordering methods with a traditional domain-based method, especially in solving large-sized problem instances.

\added[]{Although our proposed variable ordering approach is generic, we have only assessed its effectiveness on the job shop scheduling problems. To expand the scope of this research, one potential avenue is to apply this method to other constraint optimization problems. Moreover, our experiments have shown that while ordering variables based on an optimal solution ensures that the search starts with the best possible upper bound, it does not necessarily minimize the size of the search tree. Therefore, another promising direction for future research is to train a ML model with the objective of minimizing the search tree size.}

\section*{Acknowledgement}
This work was partially supported by an ARC Discovery Grant (DP180101170) from Australian Research Council.